\documentclass{egpublm}
\usepackage{egsgp2025m}
 
\SpecialIssueSubmission    %

\CGFccby

\usepackage[T1]{fontenc}
\usepackage{dfadobe}  

\usepackage{cite}  %

\usepackage{algorithm}
\usepackage{algpseudocode}
\usepackage{amsmath}
\usepackage{amssymb}
\usepackage{caption}
\usepackage{float}
\usepackage{multirow}
\usepackage{subcaption}
\BibtexOrBiblatex
\electronicVersion
\PrintedOrElectronic
\ifpdf \usepackage[pdftex]{graphicx} \pdfcompresslevel=9
\else \usepackage[dvips]{graphicx} \fi

\usepackage{egweblnk}

\title[ArchComplete]{ArchComplete: Autoregressive 3D Architectural Design Generation with Hierarchical Diffusion-Based Upsampling}

\author[S. Rasoulzadeh, M. Bank, I. Kovacic, K. Schinegger, S. Rutzinger, M. Wimmer]
{
    \parbox{\textwidth}
    {
        \centering 
        S. Rasoulzadeh $^{1}$ \orcid{0000-0002-0019-0137}, 
        M. Bank Stigsen $^{2}$ \orcid{0000-0003-3601-7907}, 
        I. Kovacic $^{1}$\orcid{0000-0002-0303-3284}, 
        K. Schinegger $^{2}$, 
        S. Rutzinger $^{2}$, 
        M. Wimmer $^{1}$\orcid{0000-0002-9370-2663}
    }
    \\
    {
        \parbox{\textwidth}
        {
            \centering 
            $^1$ TU Wien, Center for Geometry and Computational Design, Austria\\
            $^2$ University of Innsbruck, Department of Design, i.sd | Structure and Design, Austria
        }
    }
}

\begin{document}

\teaser{
 \includegraphics[width=1.0\linewidth]{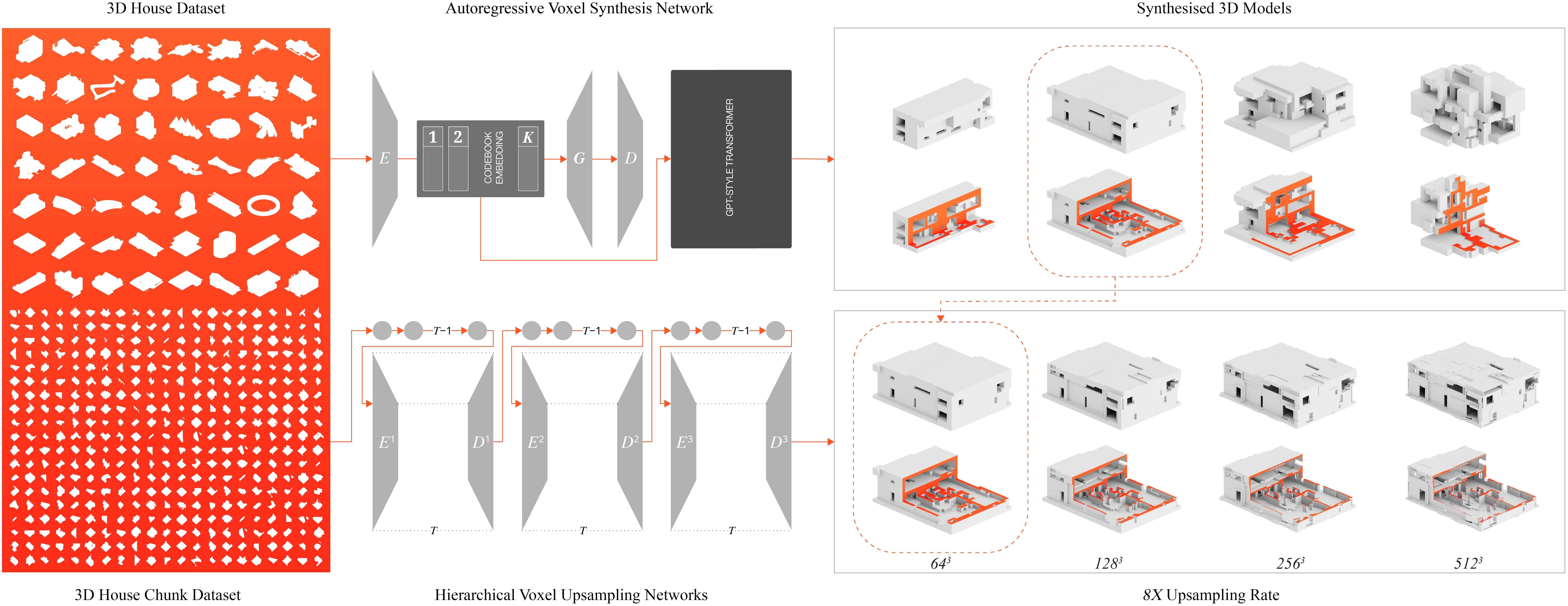}
 \centering
  \caption{\textbf{ArchComplete} synthesises novel unseen coarse 3D models and augments them with fine geometric details to streamline the design process. Trained on a dataset of 3D house models, the 3D Voxel VQGAN whose composition is modelled with a transformer, generates models at a resolution of $64^3$. Thereafter, a set of 3D conditional Denoising Diffusion Models trained on cropped local volumetric patches, i.e. chunks, progressively generate higher-resolution grids in a coarse-to-fine manner up to a resolution of $512^3$.}
\label{fig:TEASER}
}

\maketitle

\begin{abstract}
   Recent advances in 3D generative models have shown promising results but often fall short in capturing the complexity of architectural geometries and topologies and fine geometric details at high resolutions. To tackle this, we present ArchComplete, a two-stage voxel-based 3D generative pipeline consisting of a vector-quantised model, whose composition is modelled with an autoregressive transformer for generating coarse shapes, followed by a hierarchical upsampling strategy for further enrichment with fine structures and details. Key to our pipeline is (i) learning a contextually rich codebook of local patch embeddings, optimised alongside a 2.5D perceptual loss that captures global spatial correspondence of projections onto three axis-aligned orthogonal planes, and (ii) redefining upsampling as a set of conditional diffusion models learning from a hierarchy of randomly cropped coarse-to-fine local volumetric patches. Trained on our introduced dataset of 3D house models with fully modelled exterior and interior, ArchComplete autoregressively generates models at the resolution of $64^{3}$ and progressively refines them up to $512^{3}$, with voxel sizes as small as $ \approx 9\text{cm}$. ArchComplete solves a variety of tasks, including genetic interpolation and variation, unconditional synthesis, shape and plan-drawing completion, as well as geometric detailisation, while achieving state-of-the-art performance in quality, diversity, and computational efficiency. \\

    

\end{abstract}  

\section{Introduction}\label{sec:INTRODUCTION}

Fields such as architectural design rely on high-quality three-dimensional models with rich geometric details and topology, which often require a significant amount of time, compute and memory to create. Volumetric design (also known as massing or schematic design), which is usually the first and most important step in design for many, is an exemplar in this context. It often begins by constructing rough 3D shapes within a defined design space before gradually refining them to include all the details needed for exterior and interior design elements \cite{chang2021building}. However, creating a good volumetric design requires a substantial amount of time and effort. Alternatively, a designer with a rough idea of the desired shape may quickly construct a coarse shape, to which a 3D generative model then adds realistic details. Hence, such a generative pipeline capable of assisting the designer from the inception of an idea to its \textit{geometry detailisation} has the potential to greatly aid design workflows.

In recent years, there has been a surge of new and exciting work on generative models for 3D shapes. These efforts are based on various frameworks and 3D geometric representations, achieving promising results in terms of quality and diversity \cite{hui2022neural, mittal2022autosdf, cheng2023sdfusion, siddiqui2024meshgpt}. However, several challenges remain when adapting these methods to domain-specific design applications such as modelling in architecture, necessitating a tailored 3D generative pipeline addressing these challenges. First, there is a lack of detailed datasets of 3D models that encompass both exterior and interior spaces. Second, 3D models of architectural buildings/houses are often geometrically and topologically more complex, or at least significantly different from the shapes in common 3D benchmark datasets like ShapeNet \cite{chang2015shapenet} due to their structural and stylistic intricacies. Third, much of the literature on 3D generative models focuses on specific generation tasks, which cannot be easily extended to other downstream tasks (e.g., user editing), resulting in narrow application scopes. Additionally, compared to state-of-the-art 2D generative models for images, the additional third spatial dimension results in a dramatic increase in network parameters and memory-intensive feature maps in the 3D counterparts, requiring high computational resources to achieve high-resolution, detailed models. %

A 3D generative pipeline has to fulfil several criteria to accommodate the designers' needs: (i) The design produced by the pipeline must maintain a 1:1 scale relative to the training data, ensuring consistency between the spatial characteristics of the models — such as openings and room heights in the architectural domain; (ii) The generative pipeline should be capable of synthesising realistic, high-fidelity models while aiding designers in quickly exploring the design space with flexible and interactive use and refinement: For example, a designer may want to blend input models from different distinct styles to create an intermediate model (\textit{aka} shape interpolation), generate versions with varying details, or may provide partially modelled input asking for completion suggestions, i.e., shape completion. (iii) Or an artist may provide coarse shapes, thereby guiding the pipeline to create large varieties of fine and detailed geometries, a process referred to as \textit{geometry detailisation} in the remainder of this manuscript.

To address the identified challenges, we propose ArchComplete, a two-stage voxel-based 3D generative pipeline trained and tested on our introduced dataset of 3D house models featuring fully modelled exteriors and interiors. The pipeline consists of a vector-quantised generative model with an autoregressive transformer for coarse shape ideation, followed by a hierarchical upsampling strategy for augmenting shapes with fine geometric details (see Figure \ref{fig:TEASER}). In the first stage, unlike conventional 3D shape encoding schemes with continuous latent space that are generally harder to learn, we encode voxelised models into latent quantised feature maps by learning a discrete vocabulary of embeddings for \textbf{local} volumetric patches to leverage the compact and tractable discrete representation, enabling effective exploration of shape composition priors. To this end, we devise a 3D Vector-Quantised Generative Adversarial Network (\textit{3D Voxel VQGAN}) with an Autoregressive Transformer, featuring a novel 2.5D perceptual loss to ensure maintaining \textbf{global} spatial coherence and 3D PatchGAN as the discriminator to better capture geometric and topological features at the scale of local patches. We then model our 3D Voxel VQGAN's composition with a transformer that learns statistical correlations between patches, enabling autoregressive generation of local patch sequences, forming a 3D voxelised model. 
In the second stage, we aim to push the limits of purely 3D generative priors for high-resolution shape generation in dense voxel-based methods while alleviating high computational demands. To this end, we define a set of 3D conditional Denoising Diffusion Probabilistic Models (\textit{3D c-DDPMs}) \cite{ho2020denoising} that train on a hierarchy of local volumetric patches instead of entire 3D voxelised models and can upsample the coarse outputs from the first stage into finer grids. This hierarchical upsampling strategy mitigates ambiguity in voxel upsampling with large rates by introducing intermediate-level supervision while making distribution at each level easier to model, as coarse levels model rough local patches and finer levels focus on local details.

We demonstrate the versatility of our pipeline through several example applications. These include leveraging Genetic Algorithm (GA) operators to generate endless interpolations and variations of synthesised models. Additionally, we explore unconditional synthesis and two conditional synthesis tasks, shape completion and plan-drawing completion, highlighting their use cases and potential in 3D architectural design. Finally, quantitative and qualitative evidence is provided showing that ArchComplete outperforms prior methods in terms of reconstruction quality as well as both unconditional and conditional synthesis tasks based on established metrics.

Code and dataset are available at: \URL{https://gitlab.cg.tuwien.ac.at/srasoulzadeh/archcomplete.git}.

\section{Related Work}\label{sec:RELATED_WORK}

Unlike 2D images, it is less clear how to represent 3D data effectively. Various representations with pros and cons have been explored, particularly when considering 3D generative models. For instance, quite a large number of 3D generative models have been developed for point clouds \cite{achlioptas2018learning, luo2021diffusion}, dense voxel grids \cite{liu2017interactive} and more recently, sparse voxel grids \cite{schwarz2022voxgraf, ren2024xcube}, meshes \cite{siddiqui2024meshgpt}, and signed distance functions (SDFs) \cite{mittal2022autosdf, hui2022neural, cheng2023sdfusion}, etc. In this work, given our target task, we opt to use a simple and plain explicit dense voxel grid representation, as it directly corresponds to the representation of volume in 3D space, i.e. as in 3D architectural design. It shares a similar form with 2D pixels, facilitating the adoption of various image generation methods that have not been extensively explored in this domain. However, it is noteworthy that the dense nature of voxels results in increased computational resources and time requirements when generating high-resolution shapes, necessitating ingenious strategies to address these challenges.

Herein, we review state-of-the-art 3D generative and diffusion models that adapt voxel-based data representations (e.g., dense voxel grids and Truncated Signed Distance Functions (TSDFs), etc.), with a focus on their application in design.

\paragraph*{3D Generative Models.} Numerous 3D generative models build on various frameworks, including generative Variational Auto Encoders (VAEs) \cite{de2019deep, guan2020generalized, sebestyen2023using}, Generative Adversarial Networks (GANs) \cite{liu2017interactive, wang2018global, wu2022learning}, Graph Neural Networks (GNNs) \cite{zhong2023building, bauscherlearning}, and Auto Regressive (AR) models \cite{mittal2022autosdf, siddiqui2024meshgpt}. In \cite{de2019deep}, within the architectural domain, the VAE introduced in \cite{kingma2013auto} is employed for the generation, manipulation, and form-finding of building typologies represented as voxelised wireframes. Another work with architectural design use cases is \cite{sebestyen2023using}, which employs a conditional Variational Auto Encoder (c-VAE) trained on voxelised data, coupled with corresponding operative verbs (a taxonomy of simple geometrical operations) in a prototype of a generative volumetric design tool.

Our pipeline's first network is built upon \cite{van2017neural} and inspired by a later proposed method VQGAN \cite{esser2021taming}. \cite{van2017neural} first proposed a method to learn quantised and compact latent representations for 2D images using the Vector-Quantised Variational AutoEncoder (VQVAE), later followed by a version with hierarchical codebooks \cite{razavi2019generating}. VQGAN \cite{esser2021taming} learns autoregressive generation over the discrete VQVAE representations by integrating a mask generative transformer through discrete semantic codebooks. Our work utilises VQGAN's network design as its backbone and extends it to the domain of 3D voxelised shapes in architectural design. It also integrates a new loss term affecting the cohesiveness and integrity of the generated 3D models.

\paragraph*{3D Diffusion Models.} Diffusion Probabilistic Models (DPMs) \cite{ho2020denoising}, also known as Diffusion Models, have currently arisen as a powerful family of generative models. In the fields of computer graphics and vision, several recent studies have adopted diffusion models for generative 3D modelling \cite{hui2022neural, vahdat2022lion, li2023diffusion, sebestyen2023generating, shue20233d, po2024state, ren2024xcube, wu2024blockfusion}. Existing approaches mostly train a VQ-VAE on a 3D representation like voxel grids, SDFs, and Triplanes, and then employ a diffusion model in the learned latent space. NWD \cite{hui2022neural} encodes 3D shapes by building a compact wavelet representation with a pair of coarse and detail coefficient volumes through TSDF decomposition. It then formulates two networks upon DPMs to generate shapes in the form of coarse and detail coefficient volumes for generating shapes and reconstructing fine details, respectively. LION \cite{vahdat2022lion} uses a VAE framework with hierarchical DDMs in latent space that combines a global shape latent representation with a point-structured latent space, and integrates it with Shape as Priors (SAP) \cite{peng2021shape} for mesh generation. Diffusion-SDF \cite{li2023diffusion} presents a two-stage pipeline comprising a patch-wise autoencoder and a voxelised diffusion model to generate voxelised SDFs conditioned on texts. On the other hand, NFD \cite{shue20233d} is another diffusion-based 3D generative model that, instead of voxel grids, converts the occupancy field of an object in a set of axis-aligned triplane feature representations.

\section{Dataset}\label{sec:DATASET}

The current state of available 3D datasets such as BuildingNet \cite{selvaraju2021buildingnet}, Houses3K \cite{peralta2020next} or 3DBAG \cite{3dbag} presents limitations for architectural design applications. They are either inconsistent in terms of scale and labelling, or lack a consistent degree of detail with modelled interiors for generating detailed building design \cite{wang2023towards}. Comprehensive 3D datasets, especially those depicting both exterior and interior details of buildings, are therefore particularly valuable as they provide a complete representation of the spatial relations architects and designers envision.

\begin{figure}[!b]
    \centering
    \includegraphics[width=1.0\linewidth]{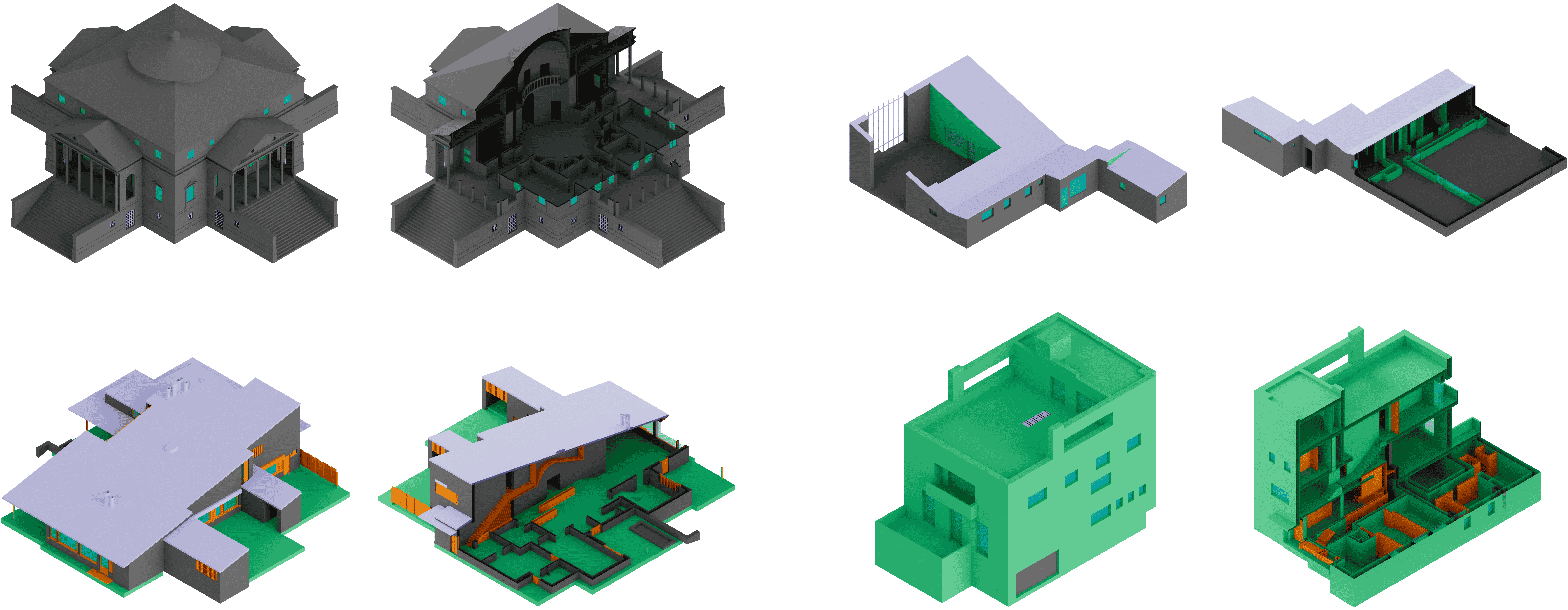}
    \caption{Example 3D houses of the dataset. The geometry is sliced with two cutting planes to reveal its corresponding interiority. Individual colours are representative of different materials.}
    \label{fig:DATASET_01}
\end{figure}

To this end, we adopt our own collected 3D House dataset, addressing the aforementioned challenge in the literature: the lack of high-quality publicly available 3D architectural datasets that capture the nuanced relationships between spatial configurations, structural integrity, and materiality — critical elements in the architectural design process. Our dataset consists of $1500$ 3D house models, including both the original and augmented samples that depict prominent architectural houses in a 1:1 scale with fully modelled interiors and material labels, while possessing a comparable level of detail with respect to each other (see Figure \ref{fig:DATASET_01}). Additionally, a key aspect of this dataset is that it consists of architectural precedents — pre-existing buildings/houses or projects that serve as inspiration for new designs. These precedents provide both explicit and implicit knowledge on how past design challenges related to spatial organisation and aesthetics \cite{cross1982designerly}. The models also only consist of solid elements such as closed poly-surfaces or meshes, with each model depicting both the exterior and interior layout of the real house with doors, windows, stairs, floors, roofs, walls and inbuilt furniture elements. 

The original samples in the dataset were collected through structured research-led teachings at the University of Innsbruck. Also, refer to the Supplementary Material for detailed information on the augmentation technique used to expand the dataset. 

\begin{figure*}[ht]
    \centering
    \includegraphics[width=1.0\textwidth]{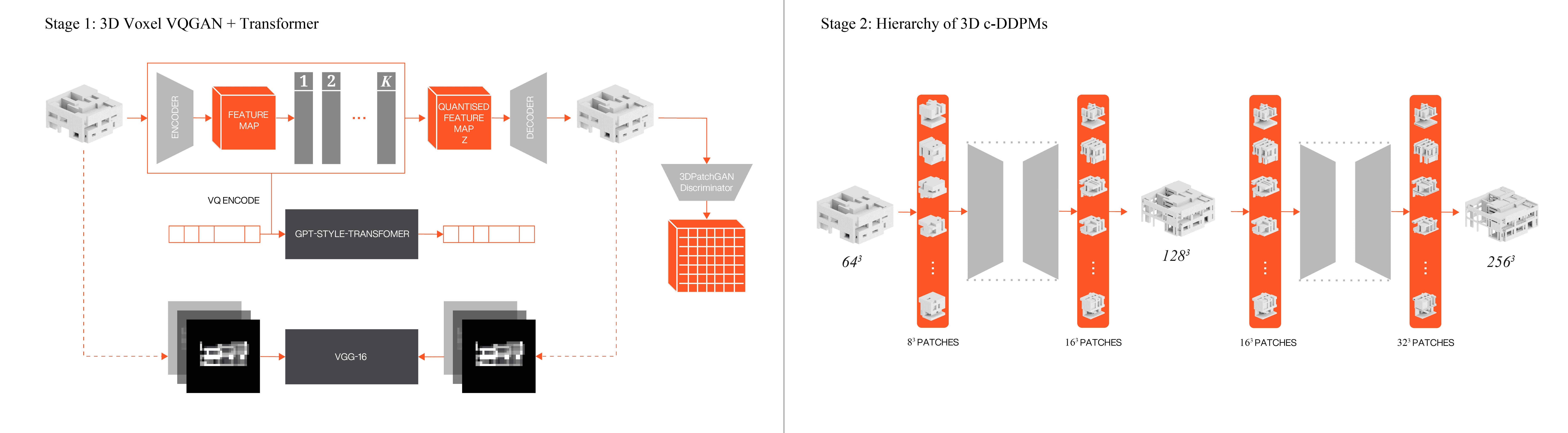}
    \caption{\textbf{Pipeline Overview}. The framework follows a two-stage 3D generative pipeline. In stage 1 (\textbf{left}), a 3D Voxel VQGAN is trained on dense voxelised data with a patch-wise quantisation step together with a 3D PatchGAN discriminator that enforces fidelity at the scale of local patches and a 2.5D perceptual loss to retain global spatial coherence. Followed by an autoregressive transformer architecture trained to synthesise new coarse voxelised models. In stage 2 (\textbf{right}), a set of 3D c-DDPMs models are trained over a hierarchy of local volumetric patches conditioned on coarser levels in a cascaded fashion. During inference, the coarse output is unfolded into overlapping patches, upsampled in parallel, and folded back into a high-resolution grid. This process continues until the highest desired resolution is achieved. Note that we achieve an $8X$ upsampling rate in our implementation, while the figure only depicts up to $4X$.}
    \label{fig:OVERVIEW_DIAGRAM}
\end{figure*}

\section{Method}\label{sec:METHOD}

Our voxel-based 3D generative pipeline can assist the designer with both early-stage design ideation and subsequent geometric detailisation of the designs. Figure \ref{fig:OVERVIEW_DIAGRAM} illustrates the overview of our proposed pipeline, named ArchComplete, which consists of two main stages: an \textit{Autoregressive Voxel Synthesis Network} for coarse shape generation in stage 1 (in Section \ref{sub_sec:STAGE_01}), followed by \textit{Hierarchical Voxel Upsampling Networks}, augmenting the coarse shape with fine geometric details in stage 2 (in Section \ref{sub_sec:STAGE_02}). Before beginning, we describe our data preparation for constructing the ground-truth data and defining the input/output for each stage using our dataset of 3D house models (in Section \ref{sub_sec:DATA_PREPARATION}).

\subsection{Data Preparation}\label{sub_sec:DATA_PREPARATION}

Prior to voxelising our data samples, as part of pre-processing, we filter out openings in the house dataset geometries, i.e., parts labelled as doors and windows, to focus solely on the mass-void relationship (closed and open spaces). Additionally, we remove parts labelled as ground to attend exclusively to the structural components of the models. We then use the dense binary voxelisation of data samples, where each voxel contains a value of either $0$ or $1$, representing an empty (void) or occupied voxel (mass), respectively. %

In order to create ground-truth for the autoregressive voxel synthesis network in the first stage, we take the models in their original 1:1 scale, limit ourselves to the design space of $48\text{m}^3$, and voxelise only the regions of the models falling within this space at a resolution of $64^3$, resulting in a voxel size as small as $75\text{cm}$. In total, we end up with $1500$ voxelised models ready for training the stage 1. 

\begin{figure}[H]
    \centering
    \includegraphics[width=0.825\linewidth]{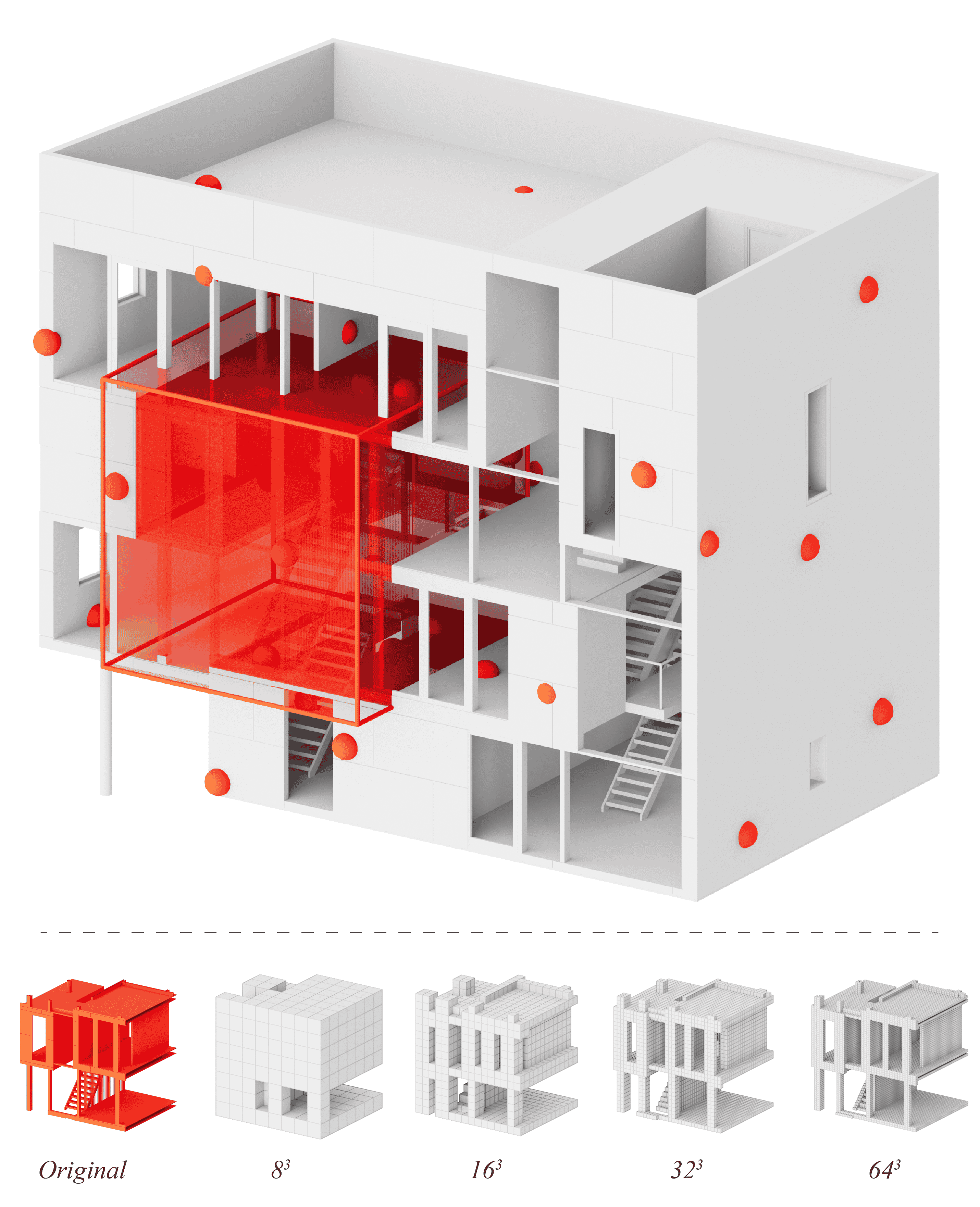}
    \caption{Sampling chunks on a 3D house model and the corresponding voxelised patches in four different resolutions forming the ground-truth data for training hierarchical upsampling networks. Note that not all $100$ points are shown for visual ease.}
    \label{fig:DATA_PREPARATION_8X} %
\end{figure}

For building the ground-truth data for the second stage, we create and define a hierarchy of local volumetric patches. To this end, we first sample $100$ approximately equidistant points using Poisson Disk Sampling \cite{yuksel2015sample} on the surface of each model from the dataset. Then, we crop $6\text{m}^{3}$ chunks centred at the sampled points, essentially coming up with another dataset which we refer to as 3D House Chunk dataset. We then voxelise the resultant chunks into four different resolutions: $8^3$, $16^3$, $32^3$, and $64^3$ as illustrated in Figure \ref{fig:DATA_PREPARATION_8X}. This results in three pairs of $150000$ coarse and fine local volumetric patches to train the stage 2.

\subsection{Stage 1: Autoregressive Voxel Synthesis Network}\label{sub_sec:STAGE_01}

In Stage 1, we propose a vector-quantised model named 3D Voxel VQGAN, which leverages the benefits of discrete geometric representations to learn a vocabulary of geometric embeddings corresponding to local patches. These embeddings are optimised using a novel 2.5D perceptual loss for global spatial coherency, enabling the models to be both encoded and decoded from them. The learned embeddings are then used to train a Transformer that captures their statistical correlations, allowing for the autoregressive synthesis of coarse voxelized models (see Figure \ref{fig:OVERVIEW_DIAGRAM}).

\subsubsection{Learning Quantised Voxel Grid Embeddings}\label{sub_sub_sec:LEARNING_QUANTISED_VOXEL_GRID_EMBEDDINGS}

Due to the formal similarity between pixels and voxels, we readily draw upon the vanilla 2D VQGAN \cite{esser2021taming} and extend it to the 3D Voxel VQGAN by making three key modifications: (i) we utilise 3D volumetric residual blocks as the backbone of its encoder $E$ and decoder $G$ so as to lift them to 3D; (ii) we replace the original discriminator with a 3D PatchGAN \cite{isola2017image} with a certain receptive field size to enforce fidelity at the scale of local volumetric patches; and (iii) we incorporate an additional 2.5D perceptual loss term to push the network to learn the global relationships between orthogonal projections in order to capture coherency of spatial features of the input data.

Given an input model, represented via a binary voxel grid $v \in \{0, 1\}^{R \times R \times R}$, we have
\begin{equation}\label{equ:EQU_01}
    z = \text{VQ}(E(v)), \hspace{0.25cm} \text{and} \hspace{0.25cm} \hat{v} = G(z),
\end{equation}
where $\text{VQ}$ is the patch-wise quantisation step which maps individual vectors of the latent feature map to their nearest vector in the discrete codebook $\mathcal{C} = \{z_{k} | z_{k} \subset \mathbb{R}^{D} \}_{k=1}^{K}$, obtaining $z \in \mathbb{R}^{r \times r \times r \times D}$ as the quantised latent feature map of $E$ with dimension $r < R$. If $r$ is too small, it lacks the capacity to represent intricate and detailed geometries. Therefore, we choose $r$ controlling the receptive field of the model to learn a specific code for non-overlapping local volumetric patches in the input.

The original vanilla VQGAN is trained by optimising a vector quantised codebook during autoencoding with the discriminator optimised to differentiate between real and reconstructed inputs. Given the reconstructed voxel grid denoted as $\hat{v}$, the overall loss of vanilla VQGAN can be written as:
\begin{equation}\label{equ:EQU_02}
    \begin{split}
        \underbrace{\lVert v - \hat{v} \rVert}_{\mathcal{L}_{\text{R}}} &+ 
        \underbrace{\lVert \text{sg}[E(v)] - z_{q} \rVert^{2}_{2} + 
        \lVert \text{sg}[z_{q}] - E(v) \rVert}_{\mathcal{L}_{\text{C}}} \\
        &+ \underbrace{[\log D(v) + \log(1 - D(\hat{v}))]}_{\mathcal{L}_\text{D}},
    \end{split}
\end{equation}
where $\text{sg[·]}$ denotes the stop-gradient operation and $\mathcal{L}_{R}$, $\mathcal{L}_{C}$, and $\mathcal{L}_{D}$ are the reconstruction, the so-called "commitment", and the discriminator losses, respectively. However, when extending to 3D voxel grids, capturing perceptually rich codebook vectors that reflect global spatial and structural relationships in complex inputs, e.g. 3D house models with exterior and interior, necessitates going beyond mere voxel-wise matching as in $\mathcal{L}_{R}$. To this end, we propose a 2.5D perceptual loss
\begin{equation}\label{equ:EQU_03}
    \mathcal{L}^{2.5}_{P} (v, \hat{v}) = \frac{\sum_{i \in \{xy, yz, xz\}} \sum_{j} \lambda_{j} \lVert \phi_{j}(f_{i}(v)) - \phi_{j}(f_{i}(\hat{v})) \rVert}{3}, 
\end{equation}
where $\phi_{j}$ represents the individual layers of a pre-trained visual perception network (in our case, VGG-16 \cite{simonyan2014very}) with a weighting factor $\lambda_{j}$, and $f_{i}$ for $i \in \{xy, yz, xz\}$ is the resultant feature maps obtained by projecting voxel grid $v$ onto three axis-aligned orthogonal planes. In the projections, features falling into the same plane grid are aggregated via spatial averaging, resulting in three feature maps corresponding to the three planes. The term $\mathcal{L}^{2.5}_{\text{P}}$ essentially captures the global spatial correspondence between the reconstructed and ground-truth grids at increasing levels of abstraction, thereby learning the coherency of the input models. Moreover, we also employ a 3D PatchGAN as our discriminator to distinguish real and reconstructed voxel grids to enforce fidelity at the scale of $R_{D} \times R_{D} \times R_{D}$ patches where $R_{D} < R$.

The final loss function $\mathcal{L}_{\text{3DVoxelVQGAN}}$ is a weighted sum of the individual loss terms:
\begin{equation}\label{equ:EQU_04}
    \mathcal{L}_{\text{3DVoxelVQGAN}} = \alpha \mathcal{L}_{R} + \beta \mathcal{L}^{2.5}_{P} + \gamma \mathcal{L}_{C} + \delta \mathcal{L}_{D}.
\end{equation}
Overall, the combination of the above terms in the loss function allows learning of both local fine-grained details and the global arrangement of voxels within the grid representing a model. 

\subsubsection{Learning the Composition of Voxel Grids with Transformers}\label{sub_sub_sec:LEARNING_COMPOSITION_OF_VOXEL_GRIDS_WITH_TRANSFORMERS}

Benefiting from a compact composition and tractable order of discrete representation, the correlation between discrete codes corresponding to local patches can be absorbedly learned, thereby effectively exploring priors of shape composition. To this end, we adopt a decoder-only transformer \cite{esser2021taming} to learn the composition of the input voxel grids.

With $E$ and $G$ available after training, if we replace each local patch's codebook vector embedding by its index in the codebook $\mathcal{C}$, we obtain a sequence $s \in \{0, \cdots K \}^{r \times r \times r}$ representing the input voxel grid $v$:
\begin{equation}\label{equ:EQU_05}
    s_{uvw} = k \hspace{0.25cm} \text{such that} \hspace{0.25cm} (s)_{uvw} = z_{k}.
\end{equation}
By mapping indices of the sequence $s$ back to their corresponding codebook entries, $z =(z_{s_{uvw}})$ can be readily recovered and decoded to a voxel grid $\hat{v} = G(z)$.

Thus, after choosing some ordering of the indices in $s$, voxel grid generation can be formulated as an autoregressive next-index prediction task. More precisely, given indices $s_{<i}$, a transformer can be learned to predict the distribution of possible following indices, i.e. $p(s_{i} | s_{< i})$ to compute the likelihood of the full representation $p(s) = \prod p(s_{i} | s_{< i})$. Essentially, this allows us to directly minimise negative log-likelihood of the data representations: 
\begin{equation}\label{equ:EQU_06}
    \mathcal{L}_{\text{3DVoxelVQGAN\_TRANS}} = \mathbb{E}_{v \sim p(v)}[- \log p(s)].
\end{equation}
where $p(x)$ is the distribution of real data. 

For training, we prefix the sequences with a Start Of Sequence (\textit{SOS}) token and employ a miniature version of the GPT model \cite{radford2018improving}. The transformer is trained to reconstruct masked sequences created by sampling a mask from a Bernoulli distribution to simulate imperfections in the token sequence. The transformer is then updated independently with the cross-entropy loss to refine sequences.

\subsubsection*{Inference}\label{sub_sub_sqc:STAGE_01_INFERENCE}

Once both the 3D Voxel VQGAN and its subsequent Transformer are trained, we synthesize 3D shapes by progressively sampling the next patch token using a top‑$k$ sampling strategy until all elements in the sequence $s$ are completed. Next, we feed the corresponding token codebook vectors into the first-stage decoder to obtain occupancy probability values for all sampled patches of size $r^{3}$, which are then folded back together to form the final synthesized grid of size $R^{3}$.

\begin{figure}[!t]
    \centering
    \begin{subfigure}[b]{0.30\columnwidth}
        \centering
        \includegraphics[width=\textwidth]{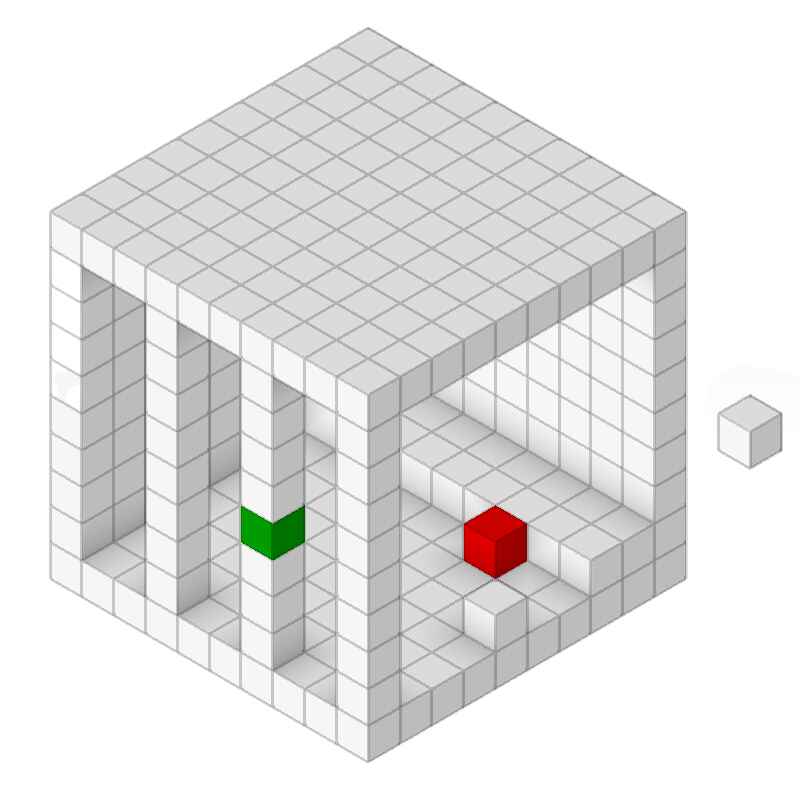}
        \caption{$V_{c} = 4$}
        \label{fig:VOXEL_CLEANUP_01}
    \end{subfigure}
    \hfill
    \begin{subfigure}[b]{0.30\columnwidth}
        \centering
        \includegraphics[width=\textwidth]{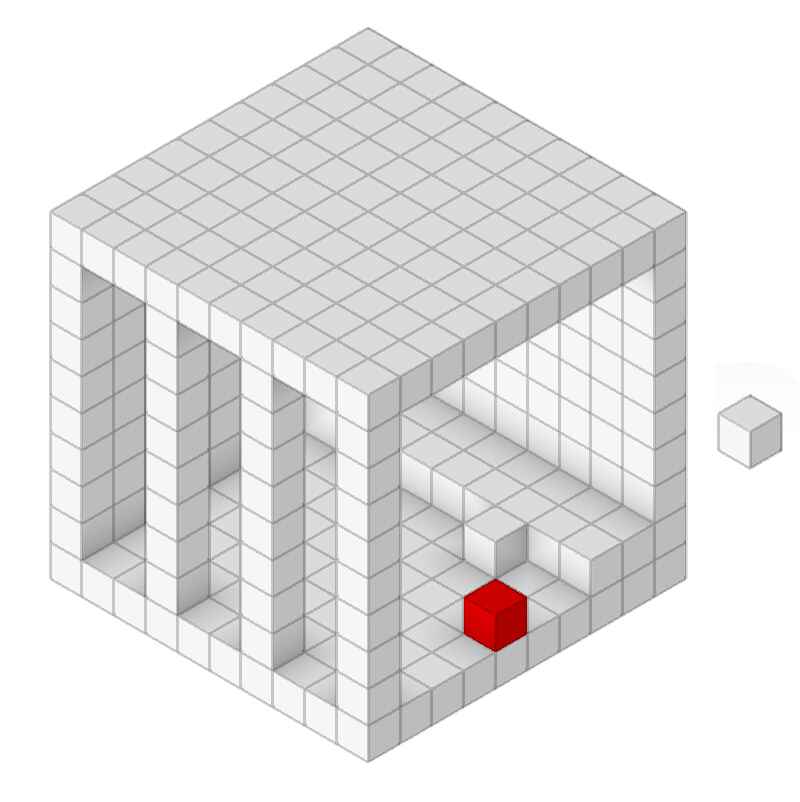}
        \caption{$V_{c} = 5$}
        \label{fig:VOXEL_CLEANUP_02}
    \end{subfigure}
    \hfill
    \begin{subfigure}[b]{0.30\columnwidth}
        \centering
        \includegraphics[width=\textwidth]{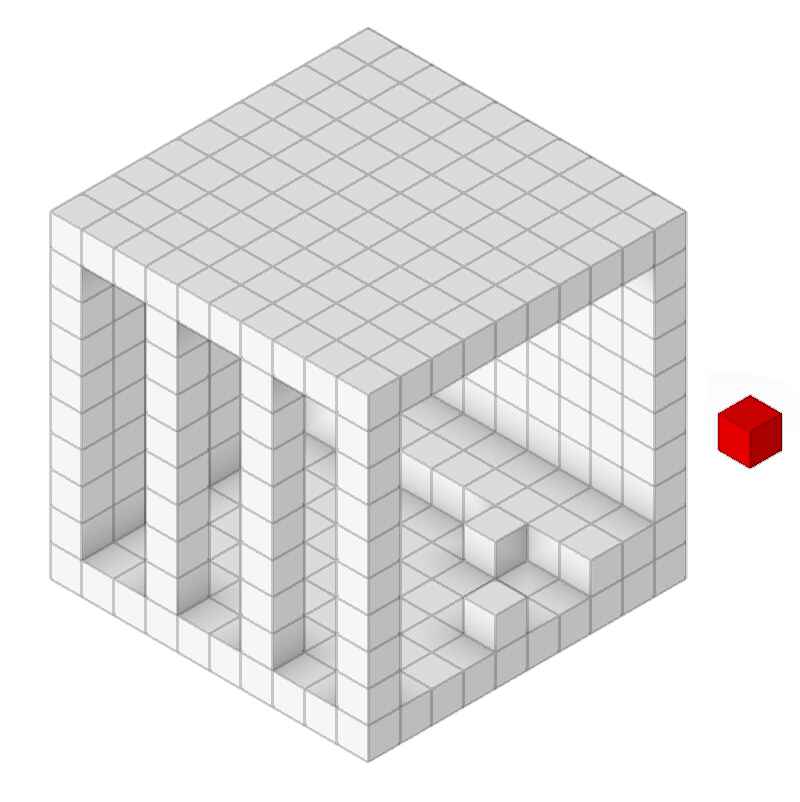}
        \caption{$V_{c} = 6$}
        \label{fig:VOXEL_CLEANUP_03}
    \end{subfigure}
    \caption{Three exemplars showcasing noisy voxel configuration. We detect such cases by computing the axis-wise $V_{c_{i}}$ and checking for their individual values and also their summation. (a) One acceptable (in green) and one noisy voxel, while (b) is a noisy sticking, and (c) is noisy floating voxel.}
    \label{fig:VOXEL_CLEANUP}
\end{figure}

\paragraph*{Regularisation.} The raw voxel grid outputs of the 3D Voxel VQGAN may contain noise artefacts, such as voxels either sticking out from the model's surface or floating in space. To clean up the voxel grid, we consider the relation and connectivity of each voxel with its face-to-face mass (1) and void (0) neighbours and devise a clean-up method by first defining the \textit{Variation Contribution} ($V_{c}$) of each voxel as the sum of the $L_1$ distances between it and its adjacent voxels along the three axes. Given this, sticking voxel have $V_c = 4$ or $V_c = 5$, while floating voxels have $V_c = 6$ (see Figure \ref{fig:VOXEL_CLEANUP}). However, while we treat voxels with the two latter scores as noise, the nuance is that voxels with  $V_{c} = 4$ are only removed if they do not belong to thin geometric structures, e.g. columns in 3D houses (see Figure \ref{fig:VOXEL_CLEANUP_01}). We detect such cases by computing the axis-wise $V_{c_{i}}$ for $i \in \{x, y, z\}$ and verifying that the multi-set $\{ V_{c_x} , V_{c_{y}}, V_{c_z} \}$ contains exactly two occurrences of 2 and one occurrence of 0. We progressively remove voxels that meet the condition while keeping the rest, resulting in a clean, noise-free voxel grid (see Algorithm \ref{alg:ALG_01} for the pseudocode).

\begin{algorithm}
    \caption{Voxel Grid Clean-Up Method}
    \label{alg:ALG_01}
    \begin{algorithmic}[1]
    \State \textbf{Input:} $v$: Input Voxel Grid, $N$: Number of iterations
    \State \textbf{Output:} $v_\text{clean}$: Output Cleaned-Up Voxel grid
    \For{$i = 1 \ \text{to} \ N$}
        \State \# Compute voxel differences:
        \State $V_{c_{x}} \gets \sum_{d=1}^{D-1}    \sum_{h=1}^{H}      \sum_{w=1}^{W}      |v_{d+1, h, w} - v_{d, h, w}|$
        \State $V_{c_{y}} \gets \sum_{d=1}^{D}      \sum_{h=1}^{H-1}    \sum_{w=1}^{W}      |v_{d, h+1, w} - v_{d, h, w}|$
        \State $V_{c_{z}} \gets \sum_{d=1}^{D}      \sum_{h=1}^{H}      \sum_{w=1}^{W-1}    |v_{d, h, w+1} - v_{d, h, w}|$
        \State $V_{c} \gets V_{c_{x}} + V_{c_{y}} + V_{c_{z}}$
        \State \# Avoid thin parts to be removed:
        \State \# $S_{n}(G)$: the group of all permutations of 
        \State \# multi-set $G$ of size $n!$
        \If{$\{V_{c_{x}}, V_{c_{y}}, V_{c_{z}}\} \in S_{3}(\{0, 2, 2\})$}
            \State $V_{c}(d,h,w) \gets 0$
        \EndIf
        \State \# Create mask $M$:
        \If{$V_{c}(d, h, w) \neq 4 \ \text{and} \ V_{c}(d,h,w) \neq 5$}
            \State $M(d,h,w) \gets 1$
        \Else
            \State $M(d,h,w) \gets 0$
        \EndIf
        \State \# Element-wise multiplication
        \State $v \gets v \odot M$
    \EndFor
    \State $v_\text{clean} \gets v$
    \State \textbf{Return} $v_\text{clean}$
    \end{algorithmic}
\end{algorithm}

\subsection{Stage 2: Hierarchical Voxel Upsampling Networks}\label{sub_sec:STAGE_02}

In stage 2, for refining low-resolution dense voxel grids from a prior stage into higher-resolution grids trains a set of 3D conditional Denoising Probabilistic Diffusion Models (3D c-DDPMs) on a hierarchy of coarse-to-fine local volumetric patches extracted from the raw dataset samples (see Figure \ref{fig:OVERVIEW_DIAGRAM}). The key idea is that coarse patches in each level of the hierarchy are to act as 3D priors to guide the generation of finer outputs so as to shift the learning focus to local structures with higher shared geometric similarity than entire 3D complex models.

Before we go deep into the details of our hierarchical voxel upsampling networks, we first provide a brief background to diffusion models. The typical denoising diffusion probabilistic generative model basically consists of a \textit{forward} and a \textit{reverse} process. During the forward phase, a sample is corrupted by adding noise in a step-wise manner, where the noise-adding process forms a Markov chain. Contrarily, the reverse phase recovers the corrupted data at each step with a denoising model \cite{lin2024discrete}. 

\textit{Forward Process}: Given a sample $x_{0} \sim q(x_{0})$, the forward Markov process $q(x_{1:T} | x_{0}) = \prod_{t} q(x_{t} | x_{t - 1})$ corrupts the sample into a sequence of increasingly noisy samples: $x_{1}, \cdots, x_{T}$, where $t$ refers to the diffusion step. Generally, the noise follows a Gaussian distribution: $q(x_{t} | x_{t-1}) = \mathcal{N}(x_{t-1}; \sqrt{1 -\beta_{t}}x_{t - 1}, \beta_{t}\textbf{I})$ where $\beta_{t}$ is the scale of the added noise at step $t$.

\textit{Reverse Process}: The reverse Markov process attempts to recover the last-step sample with a parametrised denoising model $p_{\theta}(x_{t-1} | x_{t})$. When the noise follows a Gaussian distribution, the parameterised distribution becomes
\begin{equation}\label{equ:EQU_07}
    p_{\theta}(x_{t-1} | x_{t}) = \mathcal{N}(x_{t-1} | \mu_{\theta}(x_{t}, t), \sigma_{\theta}(x_{t}, t)),
\end{equation}
where $\mu_{\theta}(x_{t}, t)$ and $\sigma_{\theta}(x_{t}, t)$ are modelled with neural networks.

The vanilla DDPM functions as an unconditional generative model, however, in our problem, our hierarchy comprises $L$ levels of coarse-to-fine patches $\mathcal{C} = \{ C^{1}, \cdots, C^{L} \}$, where each patch in the finer level $C^{l+1}$, is strictly contained within its corresponding patch in the coarser levels $C^{l}$ for $l = 1, \cdots L - 1$, with the finest patches in $\text{C}^{L}$ containing the maximum amount of detail. Coarse patches $C^{i}$ can be leveraged as conditional input for providing additional context or guidance during the reverse denoising process. To this end, we adapt 3D U-Net \cite{ronneberger2015u, cciccek20163d} as our choice of network architecture backbone to train our set of 3D conditional denoising diffusion models. We slightly modify the underlying U-Net architecture in each level \cite{ronneberger2015u} to accept three inputs — the coarse patches $C^{l}$, corresponding fine patches $C^{l +1}$, and time step $T$ — by adding an additional layer that concatenates the \textbf{subdivided} coarse and the fine voxel grid in the input layer for conditioning. The purpose of subdividing the coarse patches at each level is to align their resolution with that of their corresponding finer counterparts, thereby guiding the denoising process in the removal of extra occupied voxels to match with the ground-truth finer patches. In other words, within each level the models essentially iteratively subdivide coarse patches into octants and learn how to prune excessive ones in the denoising process.

In our 3D  conditional DDPM setup, the forward process remains almost the same, within which Gaussian noise is progressively added to the fine patches $C^{l+1}$, at times $t \in \{1, \cdots, T\}$:
\begin{equation}\label{equ:EQU_08}
    q(\text{C}_{t}^{l+1} | \text{C}_{t - 1}^{l+1}) := \mathcal{N} (\text{C}_{t}^{l+1}; \sqrt{1 - \beta_{t}} \text{C}_{t - 1}^{l+1}, \beta_{t}\textbf{I}). 
\end{equation}

Symmetrically, the reverse process modifies the Equation \ref{equ:EQU_07} by conditioning it on the coarse patches $C^{l}$ to help the model generate higher resolution patches that are consistent with the coarse ones:
{
    \small
    \begin{equation}\label{equ:EQU_09}
        p_\theta(\text{C}_{t - 1}^{l + 1} | \text{C}^{l}, \text{C}_{t}^{l + 1}) := \mathcal{N} \big(\text{C}_{t - 1}^{l + 1}; \mu_{\theta}(\text{C}_{t}^{l + 1}, t, \text{C}^{l}), \sigma_{\theta}(\text{C}_{t}^{l + 1}, t, \text{C}^{l}) \big).
    \end{equation}
}

We train the 3D c-DDPM at each level $l$ on each consecutive pair of data in our hierarchical volumetric patches $\mathcal{C}$ independently, using the following loss function:
\begin{equation}
    \mathcal{L}^{l}_{\text{3D c-DDPM}} = \mathbb{E}_{\text{C}^{l}, \text{C}^{l+1}, t, \epsilon} \Big[ \big\| \epsilon - \epsilon_\theta(\text{C}_{t}^{l+1}, t, \text{C}^l) \big\|^2 \Big],
\end{equation}
where $\epsilon$ is the Gaussian noise added to finer patches $C^{l + 1}$ during the forward diffusion process and $\epsilon_\theta(\text{C}_{t}^{l+1}, t, \text{C}^l)$ is the predicted noise by the model for the noisy patch $C^{l + 1}_{t}$ at timestep $t$, conditioned on the coarser patch $C^{l}$. 

\subsubsection*{Inference}\label{sub_sub_sqc:STAGE_02_INFERENCE}

After learning the diffusion models $p_{\theta_{l}}$ for $l = 1, \cdots, L - 1$, each model knows how to refine a given coarse coarse patch into a finer one at a timestep $t$ with an upsampling rate of $2X$. Hence, to use the diffusion models at the inference stage to upsample an entire voxel model 3D model $v$ from stage 1, we first unfold it into the patches at the coarsest scale of the hierarchy with a predefined \textbf{overlap} size. Conditioned on resultant unfolded patches $C_{v}^{1}$ of $v$, $p_{\theta_{1}}$ is then used to upsample them to $\hat{C}_{v}^{2}$. Next, an aggregation step then blends the patches by averaging the predictions from overlapping regions, ensuring smooth transitions across patches while producing the upsampled 3D model with ($2X$) resolution. The process continues until the highest resolution patches $\hat{C}_{v}^{L}$ are obtained and folded back, resulting in the final output voxelised model with  $2^{L - 1}$ times higher resolution, exhibiting finer geometric details (see Figure \ref{fig:OVERVIEW_DIAGRAM}).

\subsection{Implementation Details}\label{sub_sub_sec:IMPLEMENTATION_DETAILS}

We implement our networks using TensorFlow and run all experiments on a GPU cluster with two Nvidia A100 GPUs. The input to our 3D Voxel VQGAN consists of $R^{3} = 64^3$ voxelised model(s). In learning the codebook vectors, we set the resolution of the latent feature map $r = 8$, with the codebook constituting $K = 512$, $D = 128$-dimensional vectors. Our 3D PatchGAN discriminator has a receptive field of $R_{D} = 8$. As for the loss weight factors, we set $\alpha = 100$, $\beta = 10$, per our ablation study. We set $\lambda$ and $\delta$ to $0.25$ and $0.1$, respectively, following the vanilla 2D VQGAN \cite{esser2021taming} implementation. The base network is trained utilising the Adam optimiser \cite{kingma2014adam} with the learning rate of $10^{-4}$, while simultaneously the discriminator is trained with the learning rate of $10^{-6}$ for $128$ epochs with an effective batch size of $4$. For our transformer, we use a miniature GPT model equipped with a context window of up to $512$ embeddings and use AdamW optimiser \cite{loshchilov2017decoupled} with cosine annealing strategy \cite{loshchilov2016sgdr} with minimum and maximum learning rates of $10^{-5}$ and $2.5 \times 10^{-4}$. We train for $128$ epochs with a batch size of $32$. Finally, during inference, we set the user-defined voxel clean-up parameter $N$ (i.e., the number of iterations) to $32$.

For the hierarchical voxel upsampling networks (3D c-DDPMs), our current implementation defines a four-level hierarchy ($L = 4$), achieving an upsampling rate of $8X$. The network at each level accepts three inputs: the coarse patch(es), the corresponding fine patch(es), and the time step. It is noteworthy that we use an overlap size of one-fourth of the input patch size at each level. Furthermore, We use Adam optimiser with the learning rate of $10^{-4}$, for $256$ epochs with a batch size of $32$. For inference, to facilitate the generative sampling speed, we adopt the DDIM \cite{song2020denoising} sampler to reduce the original DDPM sampling steps from $T = 1000$ to $100$. 

For the ablation study and details of the networks' architectures, refer to the Supplementary Material.

\section{Results and Validation}\label{sec:RESULTS_AND_VALIDATION}

In this section, we first present several example applications of our pipeline (Section \ref{sub_sec:EXAMPLE_APPLICATIONS}) and subsequently evaluate various aspects of the pipeline in comparison to prior methods (Section \ref{sub_sec:EVALUATION_AND_COMPARISON}).

\subsection{Example Applications}\label{sub_sec:EXAMPLE_APPLICATIONS}

ArchComplete solves a variety of tasks. By drawing inspiration from Genetic Algorithm (GA) operators, we explore its ability to generate interpolations and variations of input models. We also demonstrate unconditional and two conditional synthesis tasks: shape completion and a specialised variant, which we refer to as plan-drawing completion as an architecture-specific use case. Finally, we showcase multi-resolution geometric detailisation capability of our pipeline. 

Refer to the Supplementary Material for more comprehensive results.

\subsubsection{Genetic Interpolation and Variation}\label{sub_sub_sec:GENETIC__INTERPOLATION_AND_VARIATION}

The quantisation step transforms an input model into a fixed-length discrete sequence of tokens corresponding to specific codebook vectors. Hence, this reframes the challenge of shape interpolation and shape variation as a well-posed problem within this discrete space, which can be addressed with genetic algorithm crossover and mutation operators to enable 3D design ideation based on existing models.

\begin{figure*}[!ht]
    \centering
    \includegraphics[width=1.0\linewidth]{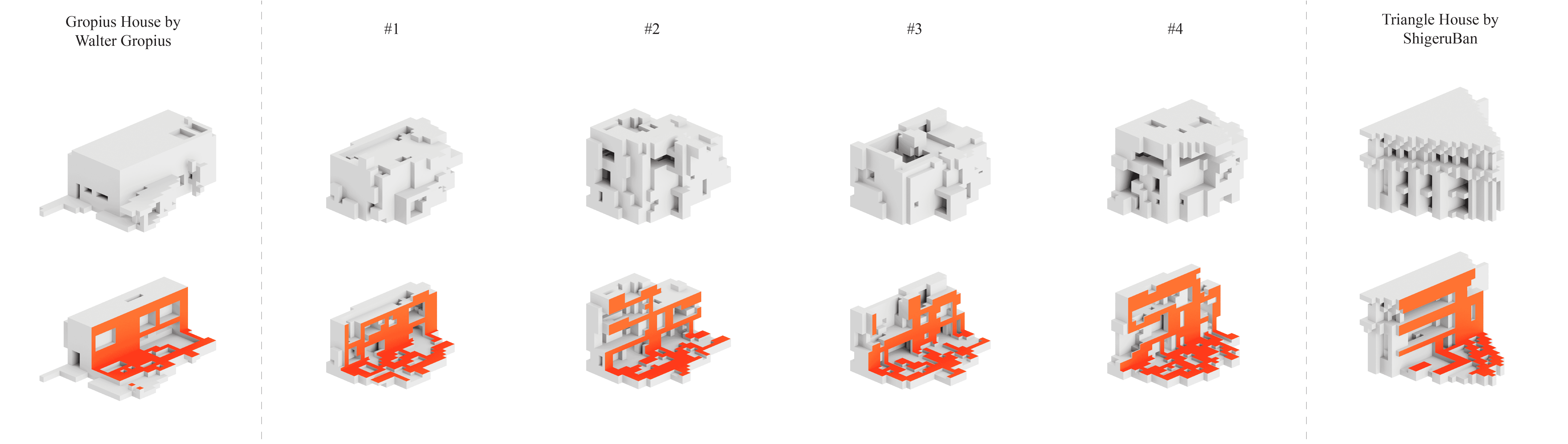}
    \caption{Four interpolated house models generated by applying crossover operations on their corresponding token sequences.}
    \label{fig:RESULTS_INTERPOLATION_01}
\end{figure*}

\begin{figure*}[!ht]
    \centering
    \includegraphics[width=1.0\linewidth]{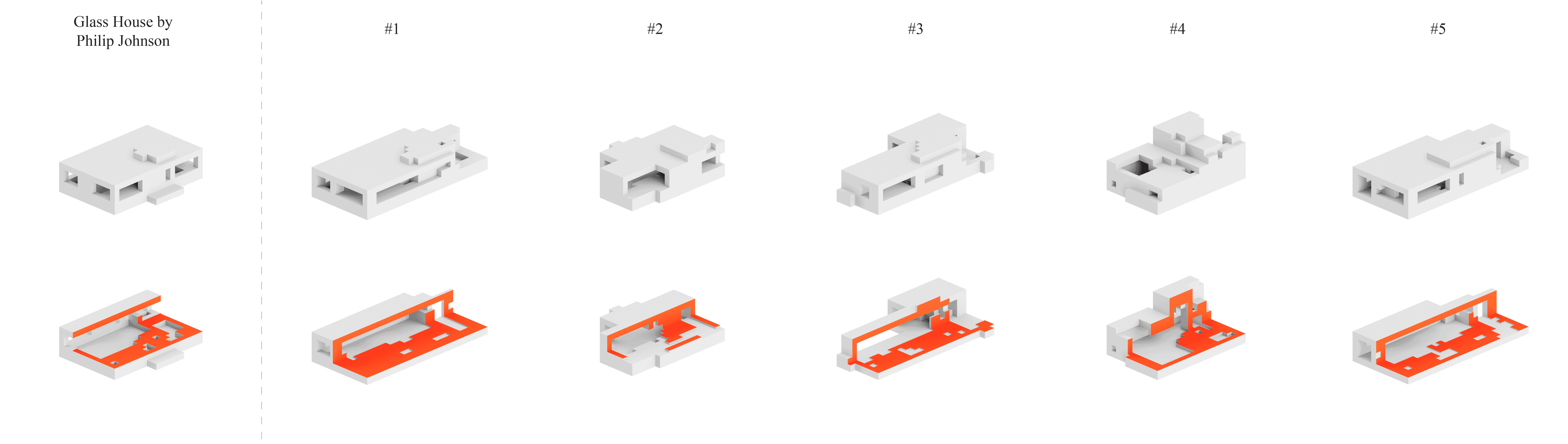}
    \caption{Five variations of an input house model generated using the swap mutation applied to codebook tokens, demonstrating how this operation creates multiple variations that retain similarities to the original 3D model while exhibit topological changes.}
    \label{fig:RESULTS_VARIATION_01}
\end{figure*}

\paragraph*{Interpolation.} Consider two codebook token sequences corresponding to the encoding of two different input models. Leveraging the random uniform crossover, we derive a new sequence by selecting each token randomly from one of the two sequences with equal probability ($50\%$). This essentially entails mixing the tokens corresponding to embeddings of local patches in the two models that produces an interpolated shape inheriting characteristics of the given two input models when decoded (see Figure \ref{fig:RESULTS_INTERPOLATION_01}).

\paragraph*{Variation.} Similarly, given a token sequence of an input model, we derive a new variant sequence using swap mutation by randomly selecting and swapping two tokens $128$ times — equal to one-fourth of the sequence length. We found this number through experimentation, introducing diversity without heavily distorting the input model. Repeating this process as many times as desired yields a corresponding number of variations of the input model upon decoding (see Figure \ref{fig:RESULTS_VARIATION_01}).

\subsubsection{Unconditional Synthesis}\label{sub_sub_sec:UNCONDITIONAL_SYNTHESIS}

\begin{figure*}[!t]
    \centering
    \includegraphics[width=1.0\linewidth]{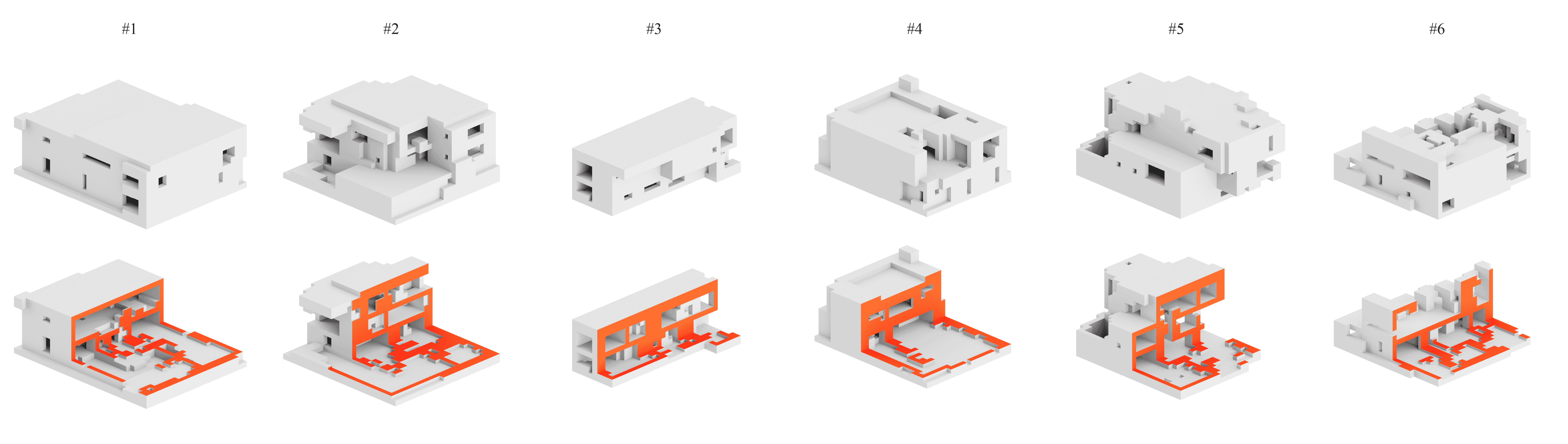}
    \caption{Six unconditionally synthesised 3D models.}
    \label{fig:RESULTS_UNCONDITIONAL_01}
\end{figure*}

With the 3D Voxel VQGAN and its subsequent transformer trained, generating novel models without prior context — \textit{aka} unconditional synthesis — is straightforward. Starting with the SOS token, we iteratively sample the next tokens using top-$k$ sampling until reaching the maximum sequence length ($512$) (see Section \ref{sub_sub_sec:LEARNING_QUANTISED_VOXEL_GRID_EMBEDDINGS} and \ref{sub_sub_sec:LEARNING_COMPOSITION_OF_VOXEL_GRIDS_WITH_TRANSFORMERS}). The sampled sequence is then decoded to synthesise a new 3D shape, demonstrating the model's ability to generate unseen diverse and coherent 3D models (see Figure \ref{fig:RESULTS_UNCONDITIONAL_01}).

\subsubsection{Conditional Synthesis}\label{sub_sub_sec:CONDITIONAL_SYNTHESIS}

In many scenarios, the designer/user demands control over the generation process by providing additional information as input from which a model shall be synthesised — \textit{aka} conditional synthesis. ArchComplete can recover the full shapes from partial inputs or a 2D plan-drawing given by the user without retraining, as our 3D Voxel VQGAN's encoder can encode information from input patches, even if some are missing. Our transformer's autoregressive structure allows us to prepend tokens for the partial input and generate completions by restricting the computation of the negative log-likelihood to remaining entries.

\begin{figure*}[!t]
    \centering
    \begin{subfigure}[H]{0.45\textwidth}
        \centering
        \includegraphics[width=\textwidth]{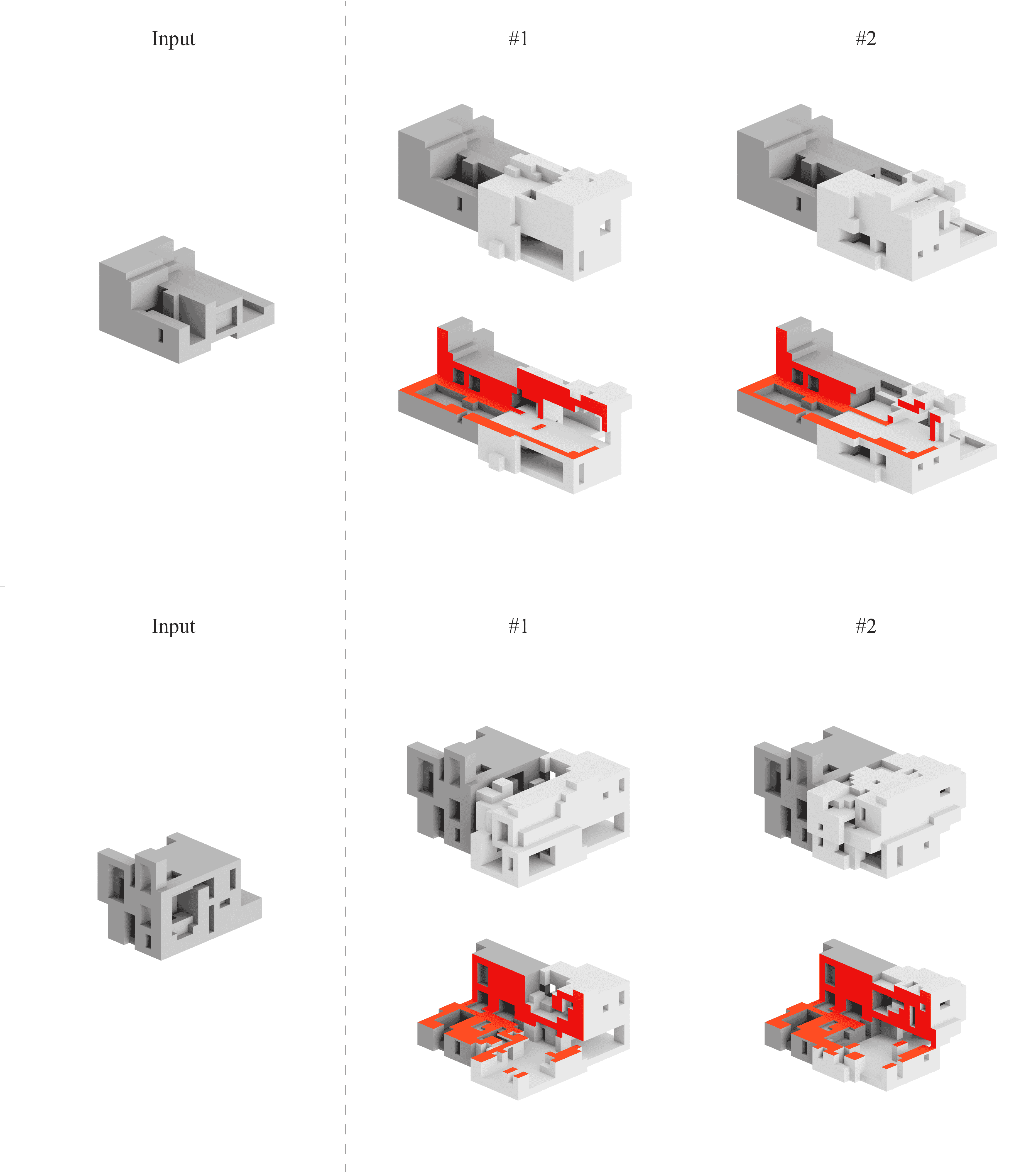}
        \caption{}
        \label{fig:RESULTS_CONDITIONAL_01}
    \end{subfigure}
    \hfill
    \begin{subfigure}[H]{0.45\textwidth}
        \centering
        \includegraphics[width=\textwidth]{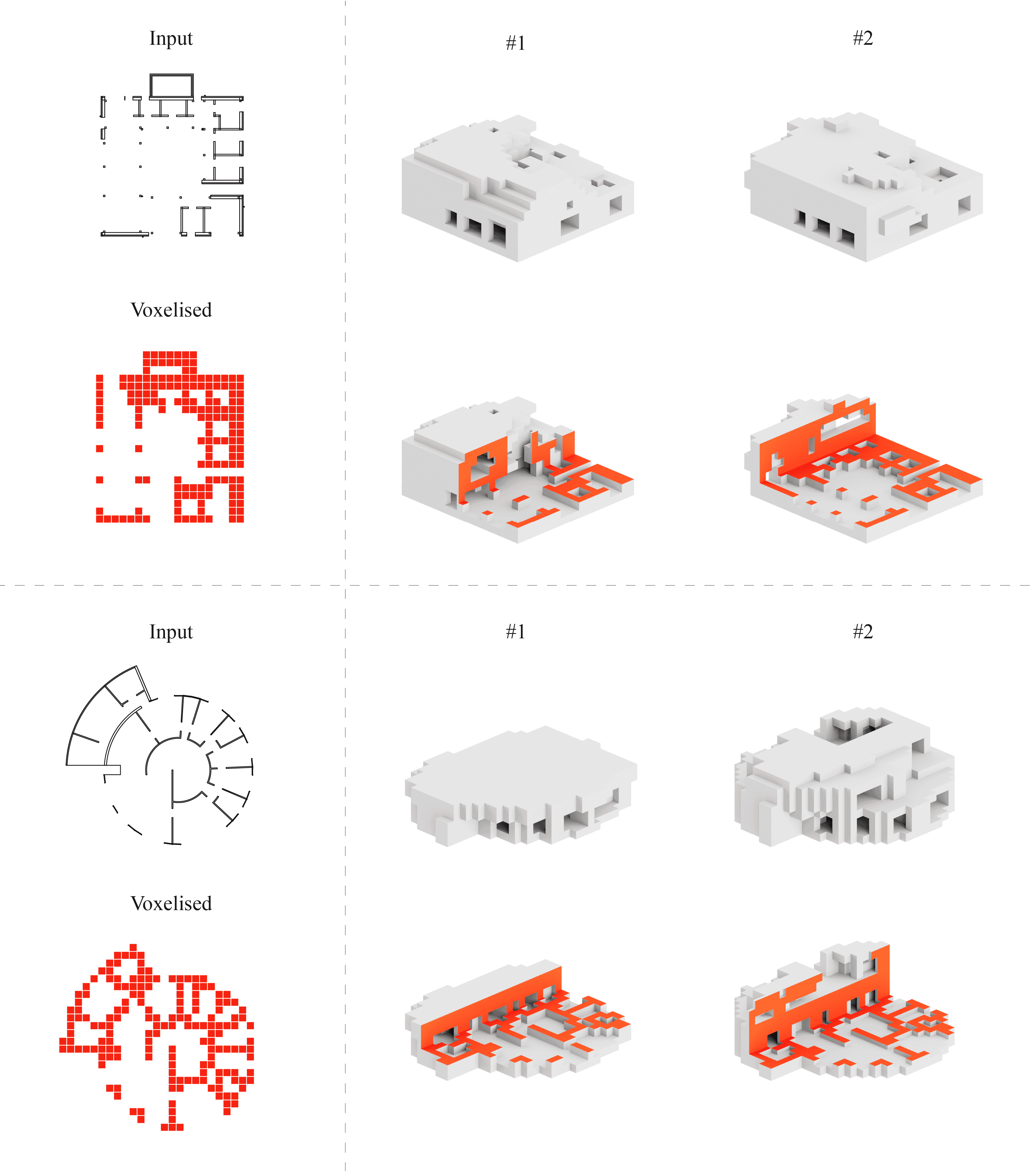}
        \caption{}
        \label{fig:RESULTS_CONDITIONAL_02}
    \end{subfigure}
    \caption{(a) ArchComplete can infer multiple possible completions for a given partial shape of a given 3D model, leveraging its probabilistic nature to generate diverse shape hypotheses that are in coordination with input. (b) Two examples of how the model can infer multiple possible completions based on a voxelised drawing of an input floor plan, highlighting how it can integrate into early stages of design.}
    \label{fig:RESULTS_CONDITIONAL}
\end{figure*}

\paragraph*{Shape Completion.} In scenarios where a designer is actively modelling, ArchComplete can suggest how the completed model might look based on partially modelled geometry as input. To illustrate the partial completion capability of ArchComplete, we create partial samples by employing a block mask on half of an unseen voxel grid of a model and use the transformer to sample the remaining tokens corresponding to patch embeddings of missing regions (see Figure \ref{fig:RESULTS_CONDITIONAL_01}).

\paragraph*{Plan-Drawing Completion.} Similarly, plan-drawing completion enables designers to create 2D plans and use ArchComplete to generate corresponding 3D models. We demonstrate this capability by asking users to create and edit their desired floor plans from a top-view on a 2D plane in Goxel \cite{chereau2023goxel}, a Minecraft-style 3D voxel editor and using the transformer to recover potential full models from the drawing (see Figure \ref{fig:RESULTS_CONDITIONAL_02}).
\subsubsection{Multi-Resolution Geometry Detailisation}\label{sub_sub_sec:MULTI_RESOLUTION_GEOMETRY_DETAILISATION}

\begin{figure*}[!t]
    \centering
    \includegraphics[width=0.95\linewidth]{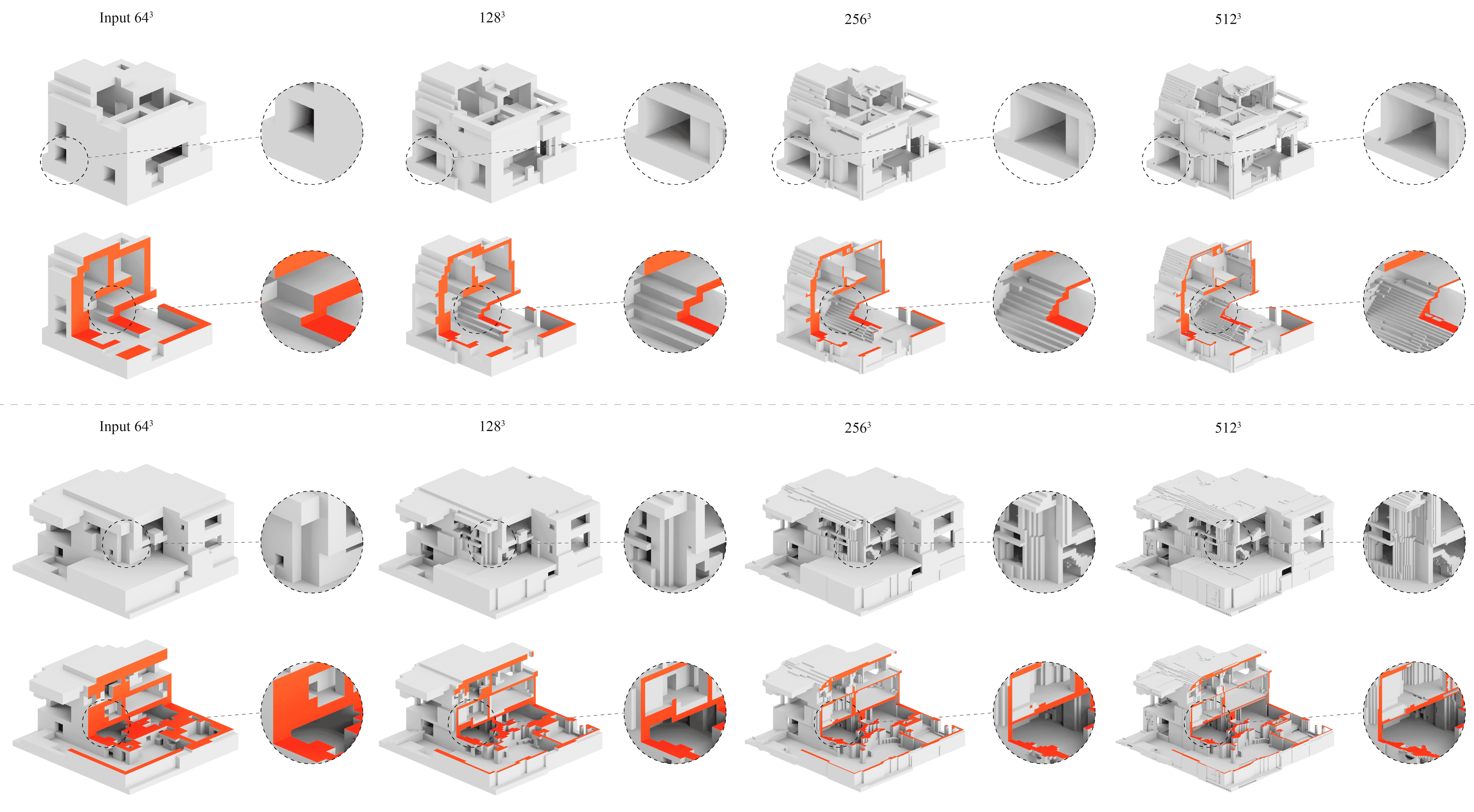}
    \caption{Four level hierarchy with $8X$ upsampling rate example results with their close-up views. The diffusion-based upsampling networks progressively augment geometries with fine structural details, refine massing, and introduce openings and elements while thinning walls and floors to achieve accurate architectural thickness.}
    \label{fig:RESULTS_UPSAMPLING}
\end{figure*}

We utilise the hierarchy of 3D c-diffusion models to incrementally upsample the initial resolution of the output models from $64^3$ to $128^3$, $256^3$, and $512^{3}$,  respectively, to augment the generated shapes with fine structures and geometric details. As ArchComplete is scale-consistent, the resolution increase reflects change in voxel size from $75\text{cm}$ to $\approx 9\text{cm}$, indicating the level of detail to which the upsampling process predicts new spatial elements and features (see Figure \ref{fig:RESULTS_UPSAMPLING}). A key advantage is a direct benefit from additional supervision at intermediate-level resolutions, which leads to higher-quality generation of finer scale geometric details.

\subsection{Evaluation and Comparison}\label{sub_sec:EVALUATION_AND_COMPARISON}

We thoroughly evaluate our network design and choice of parameters and compare our results quantitatively and qualitatively against state-of-the-art. 

We compare ArchComplete's first stage to two baselines: SDFusion \cite{cheng2023sdfusion} and NWD \cite{hui2022neural}, as well as our second stage's GPU memory usage to DECOR-GAN \cite{chen2021decor}. The first two adopt dense voxel-based data representation using TSDFs, while the latter uses the binary voxel grid representation for (high-resolution) 3D shape synthesis. 
SDFusion employs a 3D VQVAE for latent space compression, and NWD follows a two-stage approach with a coarse generator and a single-level diffusion-based detail predictor. For high-resolution shape synthesis, DECOR-GAN re-fines a coarse shape into a set of detailed shapes, each conditioned on a style code characterizing an exemplar given detailed 3D shape. Different from ArchComplete, None of the methods incorporate patch-wise encoding/upsampling, a perceptual loss to enhance coherency of generation, nor achieve our output resolution. We train both SDFusion for $128$ epochs, including its latent diffusion model for $256$ epochs, and NWD \cite{hui2022neural} for its preset epochs on our dataset using their original network architectures and parameters. It is noteworthy that for the first two methods, we extract SDF values at the resolution of $64^{3}$ for training and voxelise surface meshes at the same resolution during inference for fair comparison.

\subsubsection{Shape Novelty Analysis}\label{sub_sub_sec:SHAPE_NOVELTY_ANALYSIS}

Evaluating the unconditional synthesis of 3D shapes presents challenges due to the absence of direct ground truth correspondence. To address this, we adopt a two-fold evaluation approach: first, we assess statistically the novelty of our generated shapes with respect to the training dataset to ensure that the model is not merely retrieving existing shapes, and second, we measure our method's performance in terms of shape quality to SDFusion and NWD.

\begin{figure}[H]
    \centering
    \includegraphics[width=1.0\linewidth]{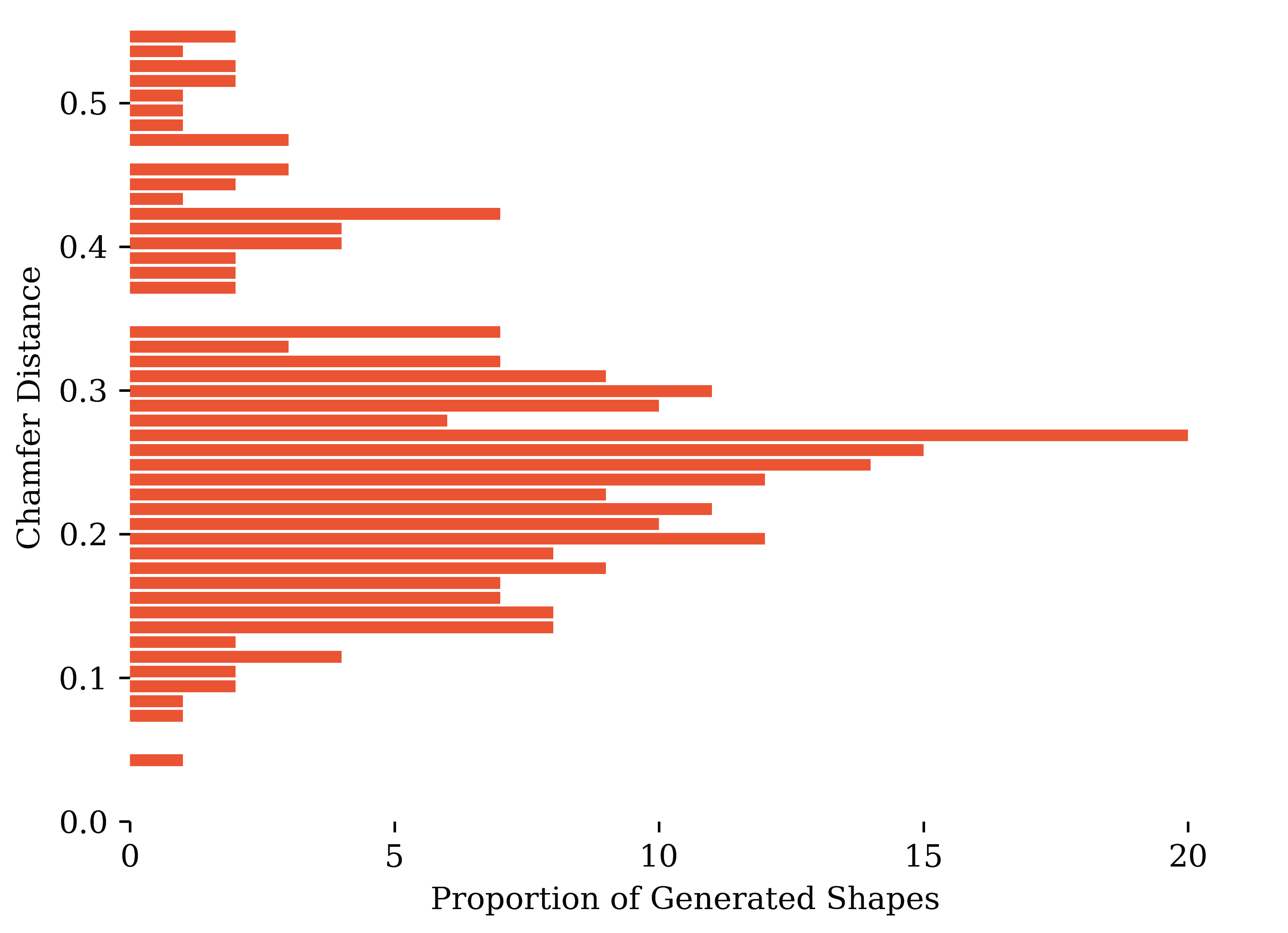}
    \caption{Shape novelty analysis on our dataset of 3D house models. The distribution of distances show that our method is able to generate shapes that are more similar (low CDs) or more novel (high CDs) compared to the training set.}
    \label{fig:EVAL_UNCONDITIONAL_01}
\end{figure}

Following the methodology in previous works \cite{erkocc2023hyperdiffusion, siddiqui2024meshgpt}, we generate $256$ shapes using and  identify the top five nearest neighbours of each from the training set based on Chamfer Distance (CD), and plot the distribution of generated house models and their closeness to training distribution (see Figure \ref{fig:EVAL_UNCONDITIONAL_01}). From the CD distribution, we can see that our method can learn a generation distribution that covers shapes in the training set and also generates novel and realistic-looking models that are more different from the training-set shapes (see Figure \ref{fig:RESULTS_UNCONDITIONAL_01}).

\begin{figure}[!ht]
    \centering
    \includegraphics[width=1\linewidth]{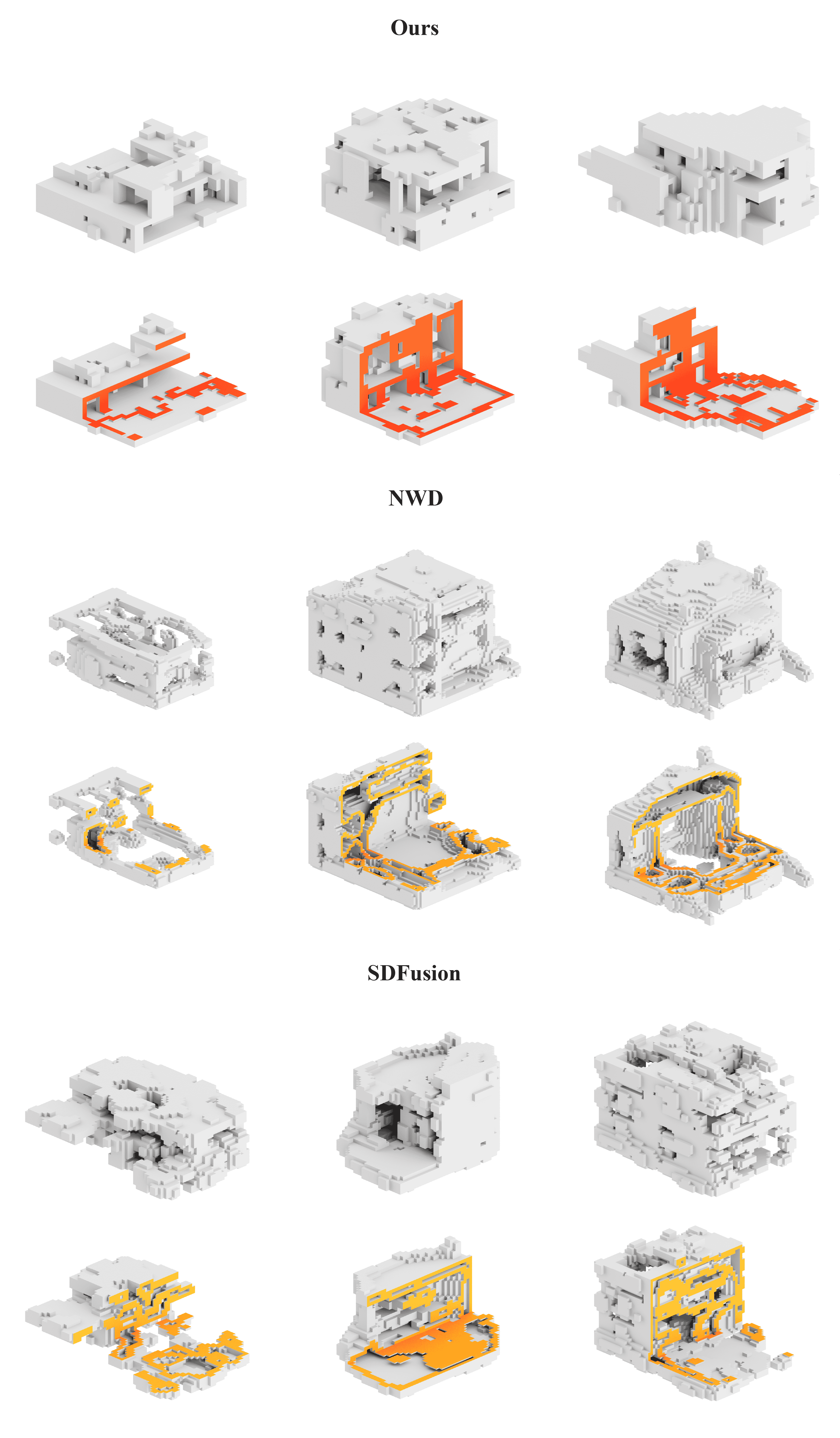}
    \caption{Qualitative comparison of the unconditional synthesis task against NWD \cite{hui2022neural} and SDFusion \cite{cheng2023sdfusion}. Compared to the baselines, our pipeline produces 3D house models with higher geometric fidelity, showcasing both exteriors and interiors with well-defined, architecturally reasonable distribution of closed and open spaces.}
    \label{fig:EVAL_UNCONDITIONAL_02}
\end{figure}

To evaluate our model's relative unconditional synthesis performance withr respect to SDFusion and NWD, we generate an equal number of shapes from each method and assess them using Coverage (COV), Minimum Matching Distance (MMD), and 1-Nearest-Neighbor Accuracy (1-NNA). As before, we compute these 3D metrics using the CD measure (see Table \ref{tab:EVAL_UNCONDITIONAL_01}). Our method outperforms both methods by generating house models with more intricate details, fewer artefacts, and architecturally sound, structured interiors of varying sizes (see Figure \ref{fig:EVAL_UNCONDITIONAL_02}). We attribute this to three main factors: first, TSDF representation is particularly effective for solid objects, however, it struggles with non-solid objects' complex interior structures and fine details at $64^{3}$ resolution; second, our tailored patch-wise encoding combined with a 2.5D perceptual loss enhances the network's ability to capture coherent structures at both local and global levels: and third, our voxel clean-up method acts as a grid smoother, producing a more continuous surface across the model.

\begin{table}[!h]
    \begin{center}
        \begin{tabular}{c|c|c|c}
            Method & COV $\uparrow$ & MMD $\downarrow$ & 1-NNA\\
            \hline
            \hline
            NWD \cite{hui2022neural}            & 35.00 & 0.85 & 81.93\\
            SDFusion \cite{cheng2023sdfusion}   & 30.50 & 0.83 & 85.54 \\
            \hline
            Ours                                & \textbf{74.50} & \textbf{0.25} & \textbf{75.51}\\
            \hline
        \end{tabular}
        \caption{Quantitative comparison on the task of unconditional synthesis. For MMD, lower is better; for COV, higher is better; for 1-NNA, 50\% is the optimal. We outperform the baselines on the shape quality metrics.}
        \label{tab:EVAL_UNCONDITIONAL_01}
    \end{center}
\end{table}

\subsubsection{Completion Fidelity and Diversity Analysis}\label{sub_sub_sec:COMPLETION_FIDELITY_AND_DIVERSITY_ANALYSIS}

We quantitatively evaluate our approach for shape completion in conditional synthesis by measuring both fidelity and diversity. Given a partial input, we generate $k$ complete shapes. For fidelity, we compute the average Unidirectional Hausdorff Distance (UHD) from the partial input to its $k$ completions. For diversity, we compute the Total Mutual Difference (TMD) as follows: given the $k$ generated results for each shape, we compute the average Chamfer Distance (CD) among the $k-1$ other shapes, and TMD is defined as the total sum of these distances. We use $k=10$ in our evaluations and report these metrics on our dataset (see Table~\ref{tab:EVAL_CONDITIONAL_01}). The quantitative results demonstrate that our completion performs favourably against the baselines, both in terms of fidelity and diversity.

\begin{table}[!h]
    \begin{center}
        \begin{tabular}{c|c|c}
            Method & UHD $\downarrow$ & TMD $\uparrow$\\
            \hline
            \hline
            SDFusion \cite{cheng2023sdfusion} & 0.0572 & 0.1252\\
            \hline
            Ours & \textbf{0.0398} & \textbf{0.2535}\\ 
            \hline
        \end{tabular}
        \caption{Quantitative comparison of conditional synthesis: shape completion results. We evaluate our method's fidelity (UHD) and diversity (TMD) in comparison to NWD \cite{hui2022neural} and SDFusion \cite{cheng2023sdfusion} trained on our dataset.}
        \label{tab:EVAL_CONDITIONAL_01}
    \end{center}
\end{table}

\clearpage

\subsubsection{GPU Memory}\label{sub_sub_sec:RUNTIME_AND_GPU_MEMORY}

Generally, one critical limitation when upsampling dense voxel grids to higher resolutions (e.g., $512^3$) is the long runtime and the severe GPU memory usage or overflow. Our proposed hierarchical upsampling networks process the input voxel grids locally (i.e., patch-by-patch), drastically reducing memory requirements when training. 

To validate our method’s GPU footprint, we adopt the generator from the upsampling network architectures used in DECOR-GAN \cite{chen2021decor} as a baseline. In their original implementation, the generator  consisting of only 3D convolutional layers is devised to upsample $X^3$ coarse voxel grids to $(4X)^3$, with a style latent code conditioning the refinement on specific geometric styles. To ensure a fair comparison, we remove the final upsampling layer to adjust the input and output resolutions to $(X)^3$ and $(2X)^3$, respectively. We also drop the conditioning layers, as they are not relevant in our evaluations. Using a batch size of $32$ samples for training, the results show that our model requires much less GPU memory, even though our networks are considerably larger in terms of number of parameters (see Table~\ref{tab:RUNTIME_GPU_MEMORY_COMPARISON_01}).

\begin{table}[!h]
    \begin{center}
        \begin{tabular}{l|c|c}
            \multirow{2}{*}{Method} & Output        & Memory \\
                                    & Resolution    & (GB)   \\
            \hline
            \hline
            DECOR-GAN & \multirow{2}{*}{$64^3 \to 128^3$} & 6.68\\
            ArchComplete &  & \textbf{1.36} \\
            \hline
            DECOR-GAN & \multirow{2}{*}{$128^3 \to 256^3$} & 35.95\\
            ArchComplete &  & \textbf{3.84}\\
            \hline
            DECOR-GAN & \multirow{2}{*}{$256^3 \to 512^3$} & OOM\\
            ArchComplete &  & \textbf{6.96}\\
            
            \hline
        \end{tabular}
        \caption{Resource computation comparison between ArchComplete and DECOR-GAN \cite{chen2021decor}. OOM: Out of Memory.}
        \label{tab:RUNTIME_GPU_MEMORY_COMPARISON_01}
    \end{center}
\end{table}

\section{Conclusion}\label{sec:CONCLUSION}

We presented ArchComplete, a voxel-based 3D generative pipeline with hierarchical diffusion-based upsampling that learns the complexity of architectural geometries and topologies, assisting with ideation and geometric detailisation in the design process. Our method frames autoregressive generation as "next local patch token" prediction, optimised with a novel 2.5D perceptual loss for global spatial coherency. Subsequently, it redefines the upsampling task by employing a cascade of conditional diffusion models trained over local patches, achieving an upsampling rate of $8X$ while reducing compute and memory requirements. In contrast to methods adopting dense voxel grids, which struggle to scale, our method unlocks unprecedented performance up to $512^3$ resolution. We further devised a novel voxel clean-up method that is capable of regularising 3D architectural geometries represented in voxel grids. Finally, we also outlined our newly collected dataset of 3D house models and results that demonstrate the ability of ArchComplete to generate complex voxelised 3D architectural models with well-defined interiors in a 1:1 scale. Extensive experiments showed that ArchComplete surpasses existing methods and establishes state-of-the-art results in the 3D generation and upsampling tasks.

\paragraph*{Limitations and Future Work.} The dataset enabled realistic generation of detailed 3D architectural models, but its limited size and diversity confine the pipeline to free-standing houses. Expanding it to include a broader range of typologies could improve the pipeline’s generalisation. An intriguing direction is to enrich the voxelised models with additional metadata (e.g., texture or material information) and further adapt the pipeline to enable semantically segmented shape generation. Combining text descriptions of individual houses in the dataset with CLIP \cite{radford2021learning} embeddings would enable image- and text-driven 3D generation, further expanding the pipeline to accept more input modalities. While the hierarchical diffusion models successfully enhance geometric detail, error accumulation is inherent in our hierarchical upsamplings, where higher-resolution grids cannot easily fix artefacts from preceding levels. This problem could potentially be mitigated by appending refinement networks to each level in the hierarchy, which we leave for future work. Another limitation of our devised geometry detailisation is that it may sometimes lead to slight topological inconsistencies in the upsampled shapes, as it lacks awareness of global structures. One possible workaround for this issue could be adding a network branch to encode global shape properties and integrating it into the latent space of the conditional diffusion models.

\bibliographystyle{eg-alpha-doi} 
\bibliography{main}

\newcommand{\etalchar}[1]{$^{#1}$}
\begin{thebibliography}{\uppercase{DMVPSC19}}

\bibitem[ADMG18]{achlioptas2018learning}
\textsc{Achlioptas P., Diamanti O., Mitliagkas I., Guibas L.}:
\newblock Learning representations and generative models for 3d point clouds.
\newblock In \emph{International conference on machine learning} (2018), PMLR, pp.~40--49.

\bibitem[BAG24]{3dbag}
\textsc{BAG D.}:
\newblock 3d bag, 2024.
\newblock Accessed: 2024-10-09.
\newblock URL: \url{https://3dbag.nl}.

\bibitem[BDEW]{bauscherlearning}
\textsc{BAUSCHER E., DAI A., ELSHANI D., WORTMANN T.}:
\newblock Learning and generating spatial concepts of modernist architecture via graph machine learning.

\bibitem[{\c{C}}AL{\etalchar{*}}16]{cciccek20163d}
\textsc{{\c{C}}i{\c{c}}ek {\"O}., Abdulkadir A., Lienkamp S.~S., Brox T., Ronneberger O.}:
\newblock 3d u-net: learning dense volumetric segmentation from sparse annotation.
\newblock In \emph{Medical Image Computing and Computer-Assisted Intervention--MICCAI 2016: 19th International Conference, Athens, Greece, October 17-21, 2016, Proceedings, Part II 19} (2016), Springer, pp.~424--432.

\bibitem[CCL{\etalchar{*}}21]{chang2021building}
\textsc{Chang K.-H., Cheng C.-Y., Luo J., Murata S., Nourbakhsh M., Tsuji Y.}:
\newblock Building-gan: Graph-conditioned architectural volumetric design generation.
\newblock In \emph{Proceedings of the IEEE/CVF international conference on computer vision} (2021), pp.~11956--11965.

\bibitem[CFG{\etalchar{*}}15]{chang2015shapenet}
\textsc{Chang A.~X., Funkhouser T., Guibas L., Hanrahan P., Huang Q., Li Z., Savarese S., Savva M., Song S., Su H., et~al.}:
\newblock Shapenet: An information-rich 3d model repository.
\newblock \emph{arXiv preprint arXiv:1512.03012} (2015).

\bibitem[Che24]{chereau2023goxel}
\textsc{Chereau G.}:
\newblock Goxel: 3d voxel editor.
\newblock \url{https://goxel.xyz/}, 2024.
\newblock Accessed: 2024-11-11.

\bibitem[CKF{\etalchar{*}}21]{chen2021decor}
\textsc{Chen Z., Kim V.~G., Fisher M., Aigerman N., Zhang H., Chaudhuri S.}:
\newblock Decor-gan: 3d shape detailization by conditional refinement.
\newblock In \emph{Proceedings of the IEEE/CVF conference on computer vision and pattern recognition} (2021), pp.~15740--15749.

\bibitem[CLT{\etalchar{*}}23]{cheng2023sdfusion}
\textsc{Cheng Y.-C., Lee H.-Y., Tulyakov S., Schwing A.~G., Gui L.-Y.}:
\newblock Sdfusion: Multimodal 3d shape completion, reconstruction, and generation.
\newblock In \emph{Proceedings of the IEEE/CVF Conference on Computer Vision and Pattern Recognition} (2023), pp.~4456--4465.

\bibitem[Cro82]{cross1982designerly}
\textsc{Cross N.}:
\newblock Designerly ways of knowing.
\newblock \emph{Design studies 3}, 4 (1982), 221--227.

\bibitem[DMVPSC19]{de2019deep}
\textsc{De~Miguel J., Villafa{\~n}e M.~E., Piskorec L., Sancho-Caparrini F.}:
\newblock Deep form finding using variational autoencoders for deep form finding of structural typologies.
\newblock In \emph{37th Conference on Education and Research in Computer Aided Architectural Design in Europe (eCAADe) \& 23rd Conference of the Iberoamerican Society Digital Graphics (SIGraDi)} (2019), eCAADe, pp.~71--80.

\bibitem[EMS{\etalchar{*}}23]{erkocc2023hyperdiffusion}
\textsc{Erko{\c{c}} Z., Ma F., Shan Q., Nie{\ss}ner M., Dai A.}:
\newblock Hyperdiffusion: Generating implicit neural fields with weight-space diffusion.
\newblock In \emph{Proceedings of the IEEE/CVF international conference on computer vision} (2023), pp.~14300--14310.

\bibitem[ERO21]{esser2021taming}
\textsc{Esser P., Rombach R., Ommer B.}:
\newblock Taming transformers for high-resolution image synthesis.
\newblock In \emph{Proceedings of the IEEE/CVF conference on computer vision and pattern recognition} (2021), pp.~12873--12883.

\bibitem[GJvK20]{guan2020generalized}
\textsc{Guan Y., Jahan T., van Kaick O.}:
\newblock Generalized autoencoder for volumetric shape generation.
\newblock In \emph{Proceedings of the IEEE/CVF Conference on Computer Vision and Pattern Recognition Workshops} (2020), pp.~268--269.

\bibitem[HJA20]{ho2020denoising}
\textsc{Ho J., Jain A., Abbeel P.}:
\newblock Denoising diffusion probabilistic models.
\newblock \emph{Advances in neural information processing systems 33} (2020), 6840--6851.

\bibitem[HLHF22]{hui2022neural}
\textsc{Hui K.-H., Li R., Hu J., Fu C.-W.}:
\newblock Neural wavelet-domain diffusion for 3d shape generation.
\newblock In \emph{SIGGRAPH Asia 2022 Conference Papers} (2022), pp.~1--9.

\bibitem[IZZE17]{isola2017image}
\textsc{Isola P., Zhu J.-Y., Zhou T., Efros A.~A.}:
\newblock Image-to-image translation with conditional adversarial networks.
\newblock In \emph{Proceedings of the IEEE conference on computer vision and pattern recognition} (2017), pp.~1125--1134.

\bibitem[Kin13]{kingma2013auto}
\textsc{Kingma D.~P.}:
\newblock Auto-encoding variational bayes.
\newblock \emph{arXiv preprint arXiv:1312.6114} (2013).

\bibitem[Kin14]{kingma2014adam}
\textsc{Kingma D.~P.}:
\newblock Adam: A method for stochastic optimization.
\newblock \emph{arXiv preprint arXiv:1412.6980} (2014).

\bibitem[LCW{\etalchar{*}}24]{lin2024discrete}
\textsc{Lin X., Chen X., Wang C., Shu H., Song L., Li B., Jiang P.}:
\newblock Discrete conditional diffusion for reranking in recommendation.
\newblock In \emph{Companion Proceedings of the ACM on Web Conference 2024} (2024), pp.~161--169.

\bibitem[LDZL23]{li2023diffusion}
\textsc{Li M., Duan Y., Zhou J., Lu J.}:
\newblock Diffusion-sdf: Text-to-shape via voxelized diffusion.
\newblock In \emph{Proceedings of the IEEE/CVF conference on computer vision and pattern recognition} (2023), pp.~12642--12651.

\bibitem[LH16]{loshchilov2016sgdr}
\textsc{Loshchilov I., Hutter F.}:
\newblock Sgdr: Stochastic gradient descent with warm restarts.
\newblock \emph{arXiv preprint arXiv:1608.03983} (2016).

\bibitem[LH21]{luo2021diffusion}
\textsc{Luo S., Hu W.}:
\newblock Diffusion probabilistic models for 3d point cloud generation.
\newblock In \emph{Proceedings of the IEEE/CVF conference on computer vision and pattern recognition} (2021), pp.~2837--2845.

\bibitem[Los17]{loshchilov2017decoupled}
\textsc{Loshchilov I.}:
\newblock Decoupled weight decay regularization.
\newblock \emph{arXiv preprint arXiv:1711.05101} (2017).

\bibitem[LYF17]{liu2017interactive}
\textsc{Liu J., Yu F., Funkhouser T.}:
\newblock Interactive 3d modeling with a generative adversarial network.
\newblock In \emph{2017 International Conference on 3D Vision (3DV)} (2017), IEEE, pp.~126--134.

\bibitem[MCST22]{mittal2022autosdf}
\textsc{Mittal P., Cheng Y.-C., Singh M., Tulsiani S.}:
\newblock Autosdf: Shape priors for 3d completion, reconstruction and generation.
\newblock In \emph{Proceedings of the IEEE/CVF Conference on Computer Vision and Pattern Recognition} (2022), pp.~306--315.

\bibitem[PCN{\etalchar{*}}20]{peralta2020next}
\textsc{Peralta D., Casimiro J., Nilles A.~M., Aguilar J.~A., Atienza R., Cajote R.}:
\newblock Next-best view policy for 3d reconstruction.
\newblock In \emph{Computer Vision--ECCV 2020 Workshops: Glasgow, UK, August 23--28, 2020, Proceedings, Part IV 16} (2020), Springer, pp.~558--573.

\bibitem[PJL{\etalchar{*}}21]{peng2021shape}
\textsc{Peng S., Jiang C., Liao Y., Niemeyer M., Pollefeys M., Geiger A.}:
\newblock Shape as points: A differentiable poisson solver.
\newblock \emph{Advances in Neural Information Processing Systems 34} (2021), 13032--13044.

\bibitem[PYG{\etalchar{*}}24]{po2024state}
\textsc{Po R., Yifan W., Golyanik V., Aberman K., Barron J.~T., Bermano A., Chan E., Dekel T., Holynski A., Kanazawa A., et~al.}:
\newblock State of the art on diffusion models for visual computing.
\newblock In \emph{Computer Graphics Forum} (2024), vol.~43, Wiley Online Library, p.~e15063.

\bibitem[Rad18]{radford2018improving}
\textsc{Radford A.}:
\newblock Improving language understanding by generative pre-training.

\bibitem[RFB15]{ronneberger2015u}
\textsc{Ronneberger O., Fischer P., Brox T.}:
\newblock U-net: Convolutional networks for biomedical image segmentation.
\newblock In \emph{Medical image computing and computer-assisted intervention--MICCAI 2015: 18th international conference, Munich, Germany, October 5-9, 2015, proceedings, part III 18} (2015), Springer, pp.~234--241.

\bibitem[RHZ{\etalchar{*}}24]{ren2024xcube}
\textsc{Ren X., Huang J., Zeng X., Museth K., Fidler S., Williams F.}:
\newblock Xcube: Large-scale 3d generative modeling using sparse voxel hierarchies.
\newblock In \emph{Proceedings of the IEEE/CVF Conference on Computer Vision and Pattern Recognition} (2024), pp.~4209--4219.

\bibitem[RKH{\etalchar{*}}21]{radford2021learning}
\textsc{Radford A., Kim J.~W., Hallacy C., Ramesh A., Goh G., Agarwal S., Sastry G., Askell A., Mishkin P., Clark J., et~al.}:
\newblock Learning transferable visual models from natural language supervision.
\newblock In \emph{International conference on machine learning} (2021), PMLR, pp.~8748--8763.

\bibitem[RVdOV19]{razavi2019generating}
\textsc{Razavi A., Van~den Oord A., Vinyals O.}:
\newblock Generating diverse high-fidelity images with vq-vae-2.
\newblock \emph{Advances in neural information processing systems 32} (2019).

\bibitem[SAA{\etalchar{*}}24]{siddiqui2024meshgpt}
\textsc{Siddiqui Y., Alliegro A., Artemov A., Tommasi T., Sirigatti D., Rosov V., Dai A., Nie{\ss}ner M.}:
\newblock Meshgpt: Generating triangle meshes with decoder-only transformers.
\newblock In \emph{Proceedings of the IEEE/CVF Conference on Computer Vision and Pattern Recognition} (2024), pp.~19615--19625.

\bibitem[SCP{\etalchar{*}}23]{shue20233d}
\textsc{Shue J.~R., Chan E.~R., Po R., Ankner Z., Wu J., Wetzstein G.}:
\newblock 3d neural field generation using triplane diffusion.
\newblock In \emph{Proceedings of the IEEE/CVF Conference on Computer Vision and Pattern Recognition} (2023), pp.~20875--20886.

\bibitem[SHR23]{sebestyen2023using}
\textsc{Sebestyen A., Hirschberg U., Rasoulzadeh S.}:
\newblock Using deep learning to generate design spaces for architecture.
\newblock \emph{International Journal of Architectural Computing 21}, 2 (2023), 337--357.

\bibitem[SME20]{song2020denoising}
\textsc{Song J., Meng C., Ermon S.}:
\newblock Denoising diffusion implicit models.
\newblock \emph{arXiv preprint arXiv:2010.02502} (2020).

\bibitem[SNL{\etalchar{*}}21]{selvaraju2021buildingnet}
\textsc{Selvaraju P., Nabail M., Loizou M., Maslioukova M., Averkiou M., Andreou A., Chaudhuri S., Kalogerakis E.}:
\newblock Buildingnet: Learning to label 3d buildings.
\newblock In \emph{Proceedings of the IEEE/CVF International Conference on Computer Vision} (2021), pp.~10397--10407.

\bibitem[S{\"O}LH23]{sebestyen2023generating}
\textsc{Sebestyen A., {\"O}zdenizci O., Legenstein R., Hirschberg U.}:
\newblock Generating conceptual architectural 3d geometries with denoising diffusion models.
\newblock In \emph{Digital Design Reconsidered-Proceedings of the 41st Conference on Education and Research in Computer Aided Architectural Design in Europe (eCAADe 2023)-Volume} (2023), vol.~2, pp.~451--460.

\bibitem[SSN{\etalchar{*}}22]{schwarz2022voxgraf}
\textsc{Schwarz K., Sauer A., Niemeyer M., Liao Y., Geiger A.}:
\newblock Voxgraf: Fast 3d-aware image synthesis with sparse voxel grids.
\newblock \emph{Advances in Neural Information Processing Systems 35} (2022), 33999--34011.

\bibitem[SZ14]{simonyan2014very}
\textsc{Simonyan K., Zisserman A.}:
\newblock Very deep convolutional networks for large-scale image recognition.
\newblock \emph{arXiv preprint arXiv:1409.1556} (2014).

\bibitem[VDOV{\etalchar{*}}17]{van2017neural}
\textsc{Van Den~Oord A., Vinyals O., et~al.}:
\newblock Neural discrete representation learning.
\newblock \emph{Advances in neural information processing systems 30} (2017).

\bibitem[VWG{\etalchar{*}}22]{vahdat2022lion}
\textsc{Vahdat A., Williams F., Gojcic Z., Litany O., Fidler S., Kreis K., et~al.}:
\newblock Lion: Latent point diffusion models for 3d shape generation.
\newblock \emph{Advances in Neural Information Processing Systems 35} (2022), 10021--10039.

\bibitem[WDL{\etalchar{*}}23]{wang2023towards}
\textsc{Wang A., Dong J., Lee L.-H., Shen J., Hui P.}:
\newblock Towards ai-architecture liberty: A comprehensive survey on designing and collaborating virtual architecture by deep learning in the metaverse.
\newblock \emph{arXiv preprint arXiv:2305.00510} (2023).

\bibitem[WLY{\etalchar{*}}24]{wu2024blockfusion}
\textsc{Wu Z., Li Y., Yan H., Shang T., Sun W., Wang S., Cui R., Liu W., Sato H., Li H., et~al.}:
\newblock Blockfusion: Expandable 3d scene generation using latent tri-plane extrapolation.
\newblock \emph{ACM Transactions on Graphics (TOG) 43}, 4 (2024), 1--17.

\bibitem[WSH{\etalchar{*}}18]{wang2018global}
\textsc{Wang H., Schor N., Hu R., Huang H., Cohen-Or D., Huang H.}:
\newblock Global-to-local generative model for 3d shapes.
\newblock \emph{ACM Transactions on Graphics (TOG) 37}, 6 (2018), 1--10.

\bibitem[WZ22]{wu2022learning}
\textsc{Wu R., Zheng C.}:
\newblock Learning to generate 3d shapes from a single example.
\newblock \emph{arXiv preprint arXiv:2208.02946} (2022).

\bibitem[Yuk15]{yuksel2015sample}
\textsc{Yuksel C.}:
\newblock Sample elimination for generating poisson disk sample sets.
\newblock In \emph{Computer Graphics Forum} (2015), vol.~34, Wiley Online Library, pp.~25--32.

\bibitem[ZKF23]{zhong2023building}
\textsc{Zhong X., Koh I., Fricker P.}:
\newblock Building-gnn: Exploring a co-design framework for generating controllable 3d building prototypes by graph and recurrent neural networks.
\newblock In \emph{International Conference on Education and Research in Computer Aided Architectural Design in Europe} (2023), eCAADe, pp.~431--440.

\end{thebibliography}

\end{document}